\documentclass{article}

\usepackage{microtype}
\usepackage{graphicx}
\usepackage{booktabs} % for professional tables

\usepackage[backref=page]{hyperref}
\usepackage{placeins}

\usepackage[accepted]{icml2025}

\usepackage{amsmath}
\usepackage{amssymb}
\usepackage{mathtools}
\usepackage{amsthm}
\usepackage{xfrac}       % compact symbols for 1/2, etc. nicefrac
\usepackage{nicefrac} 
\usepackage{bbm}

\usepackage{lipsum}

\usepackage{tikz}
\usetikzlibrary{bayesnet}
\usetikzlibrary{arrows}

\usepackage{wrapfig}
\usepackage{graphicx}          
\usepackage{subcaption}
\usepackage[capitalize,noabbrev]{cleveref}

\newcommand{\E}{\mathbb{E}}
\newcommand{\R}{\mathbb{R}}

\newcommand{\prob}{\mathbb{P}}

\newcommand{\halfmid}{\, \vert \,}

\newcommand{\defeq}{\vcentcolon=}

\newcommand{\cmark}{\ding{51}}
\newcommand{\xmark}{\ding{55}}

\newcommand{\bracket}[3]{\left#1 #3 \right#2}
\newcommand{\mbracket}[5]{\left#1 #4 \middle#2 #5 \right#3}
\renewcommand{\b}{\bracket{(}{)}}
\newcommand{\bc}{\mbracket{(}{\vert}{)}}
\newcommand{\sqb}{\bracket{[}{]}}

\newcommand{\tsum}{{\textstyle \sum}}
\newcommand{\tprod}{{\textstyle \prod}}

\DeclareMathOperator{\precision}{precision}
\DeclareMathOperator{\recall}{recall}
\DeclareMathOperator{\bernoulli}{Bernoulli}
\DeclareMathOperator{\categorical}{Categorical}

\DeclareMathOperator{\multinomial}{Multinomial}
\DeclareMathOperator{\betadist}{Beta}
\DeclareMathOperator{\uniform}{Uniform}
\DeclareMathOperator{\dirichlet}{Dirichlet}
\DeclareMathOperator{\betabin}{BetaBin}
\DeclareMathOperator{\gammadist}{Gamma}
\DeclareMathOperator{\SE}{SE}
\DeclareMathOperator{\CI}{CI}
\DeclareMathOperator{\Cov}{Cov}
\DeclareMathOperator{\Var}{Var}

\newcommand{\iid}{\overset{\scriptstyle{\text{IID}}}{\sim}}

\DeclareMathAlphabet{\mathsfit}{\encodingdefault}{\sfdefault}{m}{sl}
\SetMathAlphabet{\mathsfit}{bold}{\encodingdefault}{\sfdefault}{bx}{n}

\def\gN{{\mathcal{N}}}

\def\zHalfAlpha{z_{\nicefrac{\alpha}{2}}}

\usepackage{enumitem}
\usepackage{wrapfig}
\usepackage{bm} 

\usepackage{makecell}
\usepackage{amssymb}% http://ctan.org/pkg/amssymb
\usepackage{pifont}% http://ctan.org/pkg/pifont
\usepackage[outputdir=.,frozencache, cachedir=.]{minted} % use for generating arxiv submission
\setminted[python]{
    linenos,
    fontsize=\scriptsize,
    numbersep=4pt
}

\usepackage{tcolorbox}

\newtcolorbox[auto counter]{pbox}[2][]{
  colback=white,
  title=Snippet~\thetcbcounter: #2,
  #1,
  fonttitle= \small, %\sffamily,
  boxrule=0.5pt, 
  arc=2pt,
  top=1pt, 
  bottom=1pt,
  before={\vspace{2pt}}, 
  after={\vspace{2pt}} 
}
\def\equationautorefname#1#2\null{%
  Eq.\!#1#2\null%
}

\def\algorithmautorefname#1#2\null{
    Algorithm#1#2\null
}

\usepackage{thmtools}
\usepackage{thm-restate}
\theoremstyle{plain}

\newtheorem{remark}{Remark}

\usepackage[textsize=tiny]{todonotes}

\usepackage{upgreek}
\DeclareUnicodeCharacter{3B8}{$\uptheta$}

\icmltitlerunning{Position: Don't Use the CLT in LLM Evals With Fewer Than a Few Hundred Datapoints}

\begin{document}

\twocolumn[
\icmltitle{Position: Don't Use the CLT in LLM Evals With Fewer Than a Few Hundred Datapoints}
% Don't use CLT-Based Methods for Uncertainty Quantification in Language Model Evaluations

%\icmltitle{The Bayesian approach to large language models evaluations}
% Hierarchical Bayesian Model of Language Models Evaluations



\icmlsetsymbol{equal}{*}

\begin{icmlauthorlist}
\icmlauthor{Sam Bowyer}{bristol}
\icmlauthor{Laurence Aitchison}{equal,bristol}
\icmlauthor{Desi R. Ivanova}{equal,oxford}
\end{icmlauthorlist}

\icmlaffiliation{bristol}{University of Bristol, UK}
\icmlaffiliation{oxford}{University of Oxford, UK}

\icmlcorrespondingauthor{Sam Bowyer}{sam.bowyer@bristol.ac.uk}
\icmlcorrespondingauthor{Laurence Aitchison}{laurence.aitchison@gmail.com}
\icmlcorrespondingauthor{Desi R. Ivanova}{desi.ivanova@stats.ox.ac.uk}

\icmlkeywords{Machine Learning, ICML}

\vskip 0.3in
]



\printAffiliationsAndNotice{\icmlEqualContribution} % otherwise use the standard text.

\begin{abstract}
  % The Abstract must identify the paper as a position paper and briefly state the position
  % It has been recently been argued (Miller  2024, Hermann et al. 2024, Madaan et al. 2024), that rigorous statistical evaluation, including error bars and model comparison is necessary for evaluating LLMs.
  % However, these error bars and model comparison methods are typically built on the Central Limit Theorem (CLT).
  % Here, we show that CLT based statistical methods are appropriate for LLM benchmarks with more than a few hundred questions.
  % However, LLMs are increasingly evaluated on benchmarks where each question is very complex and difficult and as such, fewer questions are available.
  % In these settings, we show that CLT based statistical methods perform very poorly, usually dramatically underestimating the uncertainty (i.e.\ error bars are too small).
  % We give recommendations on more appropriate frequentist and Bayesian statistical methods.
%
Rigorous statistical evaluations of large language models (LLMs), including valid error bars and significance testing, are essential for meaningful and reliable performance assessment.
Currently, when such statistical measures are reported, they typically rely on the Central Limit Theorem (CLT).
In this position paper, we argue that while CLT-based methods for uncertainty quantification are appropriate when benchmarks consist of thousands of examples, they fail to provide adequate uncertainty estimates for LLM evaluations that rely on smaller, highly specialized benchmarks. 
In these small-data settings, we demonstrate that CLT-based methods perform very poorly, usually dramatically underestimating uncertainty (i.e.\ producing error bars that are too small).
We give recommendations for alternative frequentist and Bayesian methods that are both easy to implement and more appropriate in these increasingly common scenarios.
We provide a simple Python library for these Bayesian methods at \url{https://github.com/sambowyer/bayes_evals}.
\end{abstract}

\section{Introduction} \label{sec:introduction}

% from the trenches paper: ''While we believe more rigorous and widespread statistical testing in LM evaluation is still needed, we hope that this will spur the community to report and be more aware of statistical significance concerns and lower the difficulty of reporting such measures.

Benchmarks provide a systematic way for measuring the capabilities and risks of large language models (LLMs), for tracking progress over time as well as enabling performance comparison across different models.
Such language model evaluations (``LLM evals'') inform critical decisions about model selection and deployment.
However, current benchmarking practices rarely quantify the inherent statistical uncertainty in these evals, which  can substantially undermine the validity of and confidence in the reported results \citep{marie2021scientific,reuel2024betterbench,biderman2024lessons}.

Recent works have recognized the importance of statistical rigour in LLM evals and the need to improve it, for instance, through the inclusion of error bars \citep{dror2018hitchhikers,miller2024adding,madaan2024quantifying,hermann2024experimental}. 
When such uncertainty estimates are reported at all, they are most often asymptotic, based on Central Limit Theorem (CLT).

%
% While there is growing recognition of the need for more informative reporting \citep{dror2018hitchhikers,miller2024adding,madaan2024quantifying,hermann2024experimental}, for instance through the inclusion of error bars, such uncertainty quantification is still largely missing in practice.
%error bars are still largely missing.
%Recently, there has been increasing push towards reporting not just point estimates of performance but also measures of uncertainty, such as error bars, to better convey the reliability of these evaluations \cite{
%
In this position paper, we take it as read that LLM evals should come with error bars.
Instead, we ask about the best way to compute those error bars.
% In particular, 
% the most commonly used approach relies on frequentist methods and the Central Limit Theorem (CLT) \citep{miller2024adding,dubey2024llama3,madaan2024quantifying}.
% \textbf{Here, we argue that while the CLT works well in LLM evals with thousands of examples, it may become problematic especially we move to benchmarks where each question is more complex, and as such, fewer examples are available.  In these settings, we argue that accurate uncertainties require us to use non-CLT-based frequentist or Bayesian methods that take account of the discreteness of the outcomes.}
\textbf{Here, we argue that while CLT-based methods work well in LLM evals with thousands of examples, they systematically fail to provide valid uncertainties for smaller, specialized benchmarks, which are becoming increasingly common.
In these settings, %we argue that 
accurate uncertainty quantification requires more appropriate frequentist or Bayesian methods.
}

Many prominent LLM benchmarks such as Big Bench~\citep{srivastava2022beyond}, MMLU~\citep{hendrycks2020measuring}, GSM8K~\citep{cobbe2021training}, have large evaluation sets, on the order of hundreds to thousands.
Such benchmarks focus on relatively straightforward tasks that many LLMs have largely saturated (e.g. high school science questions), and generally do not accurately represent the tasks LLMs are used for in practical real-world applications \citep{raji2021ai,hardy2024more}. 
In contrast, both industry practitioners developing proprietary benchmarks, and researchers evaluating advanced capabilities of frontier models, such as advanced reasoning, multi-turn tool use or tasks involving specialized domain expertise, increasingly focus on  more targeted and representative benchmarks with very high quality labels. %(which are also less susceptible to leakage into training data). 
These benchmarks are much more expensive to construct and therefore tend to involve far fewer examples, often on the order of tens to hundreds per task. 

There are plenty of examples that illustrate this trend.
For instance, CUAD \citep{hendrycks2021cuad} is a specialized contract understanding dataset that consists of 510 labelled documents across 25 contract types. 
It required extensive annotation efforts from law students and reviews from experienced attorneys, with an estimated cost of around \$2 million.
FrontierMath \citep{glazer2024frontiermath,time2024tests}---a  new benchmark developed through a collaboration with over 60 expert mathematicians---contains around 300 problems across 23 categories, some of which have fewer than 3 samples.
Other benchmarks relevant to today's frontier models include \citet{aime} with 15 competition math problems, SWE Bench Verified, containing 500 samples across 4 difficulty levels \citep{jimenez2024swebench}, MLE Bench with 75 Kaggle competitions~\cite{chan2024mle}, and LiveBench, which frequently updates tasks across 6 categories, %like math, coding and reasoning, and averages 
currently averaging  55 examples per task \cite{white2024livebench}.

% As data from two different sub-tasks cannot possibly be considered independent and identically distributed (IID), you cannot naively use the size of the overall dataset to claim that the CLT is applicable.
% Instead, you need to reason about each sub-task separately, and as each sub-task is often far smaller, you again can face problems with a CLT-based approach.
Furthermore, even large benchmarks such as Big Bench are often broken down into smaller sub-tasks.
As data from two different sub-tasks cannot possibly be considered independent and identically distributed (IID), we cannot naively use the size of the overall dataset to justify applying the CLT.
Instead, each sub-task should be treated separately, and because these sub-tasks are typically much smaller, CLT-based approaches again become unreliable.

% In particular, we focus on settings with simulated data, to ensure that the right answer is known, so that we can for instance evaluate the probability that our confidence interval captures the right answer by simulating many datasets.

% To ensure that the ground truth is known, we run a large suite of experiments with simulated data, construct various interval estimators, and measure their coverage---how often the constructed confidence intervals capture that true value.
In this paper, we show that CLT-based confidence intervals can be extremely problematic in these small data regimes.
To make robust comparisons against alternative methods, both frequentist and Bayesian, we conduct a large suite of experiments with realistic simulated data, where the true parameter values are known.
We  construct various intervals and measure their \emph{coverage}, that is, the proportion of times those intervals contain the ground truth.
We find that, particularly with smaller datasets, the CLT  produces unreliable intervals that fail to achieve their target confidence level, also known as \emph{nominal coverage}. %dramatically underestimate uncertainty.
This is true across a variety of settings:  when evaluating a single model and when comparing two models, both with IID and clustered  questions.
%This is true in various settings. 
% For a single model, we consider a standard setting with a single benchmark, along with benchmarks with clustered questions, where each cluster consists of multiple reading comprehension questions based on a single passage of text.
% For model comparison, we consider various settings,
% in settings where the answers of each model are independent or correlated.

% First, when evaluating a single model, we consider the standard simple setting where all questions in the benchmark are assumed to be IID. 
% We also consider clustered questions, where each cluster consists of multiple reading comprehension questions based on a single passage of text.
% Second, when comparing models, we consider scenarios where model answers are independent or correlated.
%
In  simpler settings, such as computing error bars for a single model on IID data,  
we find that there are more appropriate, non-CLT-based frequentist methods that perform as well as Bayesian methods. 
However, in more complex settings, such frequentist methods are less readily available, while Bayesian methods allow us to easily %make the required extensions, including:
extend to cases such as 
clustered data  (e.g. the reading comprehension task discussed above) or when dealing with arbitrary metrics that are not averages of IID variables (e.g. $F$-scores).

\begin{figure}[t]
    \centering
    \includegraphics[width=0.90\columnwidth]{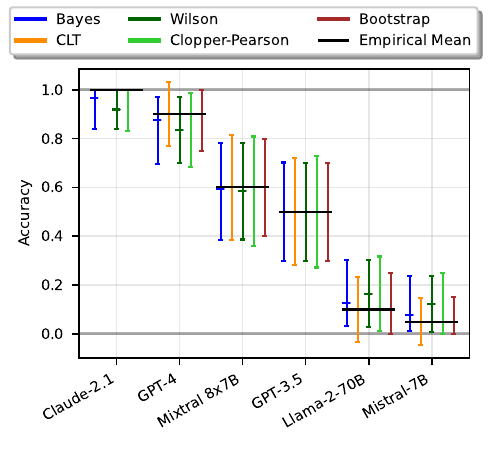}
    \vspace{-15pt}
    \caption{
        \textbf{Error bars on LangChain tool-use benchmark.} 95\% intervals for model accuracy on $N{=}20$ questions.
        The CLT produces invalid intervals, extending beyond  $[0,1]$ or collapsing to zero, highlighting its unreliability in practical settings.
        The alternative frequentist and Bayesian methods we advocate yield valid and well-calibrated intervals even in this small-data regime.
        See \autoref{app:real_data} for more  results. 
    } \label{fig:real_data_langchain}
\vspace{-10pt}
\end{figure}

% This is a terrible idea - a lot of the p hacking etc start there.
% \vspace{-10pt}
\section{The Central Limit Theorem and  Frequentist Uncertainty Quantification}
\label{sec:clt}
% \vspace{-5pt}

The Central Limit Theorem is a foundational result in frequentist statistical inference.
It states that, for sufficiently large sample size $N$, the sampling distribution of the sample mean will be approximately normal, regardless of the distribution of the original population.

Formally, if $X_1, \dots, X_N$ are \textbf{independent and identically distributed} random variables with mean $\mu \in \R$ and finite variance $\sigma^2 > 0$, then 
\vspace{-2pt}
\begin{equation}
    \sqrt{N} (\hat{\mu} - \mu) \xrightarrow{d} \gN \left( 0, \sigma^2 \right) \; \text{as } N \rightarrow \infty, \label{eq:CLT}
\vspace{-2pt}
\end{equation}
where $\hat{\mu} = \frac{1}{N}\sum_{i=1}^N X_i$ is the sample mean. Put differently, the distribution of $\hat{\mu}$ is arbitrarily close to $\gN(\mu, \frac{\sigma^2}{N})$ for~large enough $N$.
%\footnote{
The result extends verbatim to the multivariate setting: if each $X_i$ is a $p$-dimensional random vector with mean $\mu$ and covariance $\Sigma$, then $\hat{\mu} = \frac{1}{N}\sum_{i=1}^N X_i$ is approximately $\gN(\mu, \frac{\Sigma}{N})$ then for large $N$ \citep{van2000asymptotic}.
%}

The CLT underlies many common frequentist methods for uncertainty quantification, such as confidence intervals and  hypothesis tests. 
To apply it in practice, however, we need to know population variance $\sigma^2$. 
Of course we rarely know $\sigma^2$ and so we estimate it empirically by the sample variance,
    $S^2 = \frac{1}{N-1}\sum_{i=1}^N (X_i - \hat{\mu})^2$.
By Slutsky's theorem, replacing $\sigma^2$ with its \emph{consistent} estimator $S^2$ in \autoref{eq:CLT} preserves the asymptotic normality \citep[see e.g. Theorem~11.3.2 in]
[]{lehmann1986testing}, so that we have  $\sqrt{N}(\hat{\mu} - \mu) \approx \gN(0, S^2)$. %\citep[e.g. Thm~11.3.2 in][]{lehmann1986testing}.

The standard error of the sample mean, $\SE(\hat{\mu})$, is the standard deviation of the sampling distribution of $\hat{\mu}$, and equals $\SE(\hat{\mu}) = \sqrt{\nicefrac{S^2}{N}}$.

% \vspace{-10pt}
\subsection{CLT-Based Confidence Intervals} \label{sec:clt_ci}
% \vspace{-5pt}

For a desired confidence level $1-\alpha$, typically 95\% ($\alpha=0.05$) or 99\% ($\alpha=0.01$), the general form of a two-sided CLT-based confidence interval (CI) for the mean  $\mu$ is
\vspace{-2pt}
\begin{equation}
    \CI_{1-\alpha}(\mu) = \hat{\mu} \pm z_{\nicefrac{\alpha}{2}}\,\SE(\hat{\mu}), \label{eq:base_CI}
\end{equation}
where $z_{\nicefrac{\alpha}{2}}$ is the $\nicefrac{\alpha}{2}$-th quantile of the standard normal distribution (e.g. $z_{0.025} \approx 1.96$ for a 95\% CI).

\textbf{Two independent samples~~} 
Often we want to compare the means $\mu_A$ and $\mu_B$ from two independent samples with sizes $N_A$ and $N_B$ and sample variances  $S_A$ and $S_B$. The parameter of interest is then usually their difference $\mu_A - \mu_B$.
Under the independence assumption, the variance of the difference in sample means is the sum of their individual variances, so the standard error of the difference is
    $\SE(\hat{\mu}_A - \hat{\mu}_B) = \sqrt{\nicefrac{S_A^2}{N_A} + \nicefrac{S_B^2}{N_B}}$.
The $1-\alpha$ confidence interval for $\mu_A - \mu_B$ then follows the same template as \autoref{eq:base_CI}:
\begin{equation}
    \CI_{1-\alpha}(\mu_A - \mu_B) 
    = (\hat{\mu}_A - \hat{\mu}_B)  \pm z_{\nicefrac{\alpha}{2}}\,\SE(\hat{\mu}_A - \hat{\mu}_B).\label{eq:2indep_sample_clt_ci}
\end{equation}
\textbf{Two paired samples~~} 
In many scenarios, the two samples are \textit{paired}, meaning each observation in sample $A$ is naturally matched with one in sample $B$. 
Let $(X_{A,i}, X_{B,i})$ for $i=1,\ldots,N$ and define the difference $D_i = X_{A,i} - X_{B,i}$. 
The parameter of interest for which we wish to construct a confidence interval is $\mu_D = \E[D_i]$.
This setup arises often,  for example in \emph{before-after} studies where measurements are taken on the same subjects before and after a treatment or models are evaluated before and after interventions such as RLHF.
The CLT applies directly to the differences $D_1, \dots, D_N$, which we treat as a single sample and the form of the confidence interval the same as in \autoref{eq:base_CI}:%, i.e.
\vspace{-5pt}
\begin{equation}
    \CI_{1-\alpha}(\mu_D) = \hat{\mu}_D \pm z_{\nicefrac{\alpha}{2}}\,\SE(\hat{\mu}_D).\label{eq:paired_clt_ci}
    \vspace{-5pt}
\end{equation}

% \vspace{-10pt}
\subsection{CLT-Based Hypothesis Testing}\label{sec:clt_nhst}
% \vspace{-5pt}
There is a direct correspondence between null hypothesis significance testing (NHST) and confidence intervals: for a two-sided test at significance level $\alpha$, we reject $H_0$ if and only if the $1-\alpha$ confidence interval not contain zero.
The relationship extends to one-sided tests with appropriate modifications \citep[see \S\!~3.5, 5.4 in][]{lehmann1986testing}. 

To illustrate this connection, consider testing whether a population mean $\mu$ equals some hypothesized value $\mu_0$. The classical setup defines a null hypothesis $H_0: \mu = \mu_0$ and an alternative, $H_1: \mu \neq \mu_0$ (or a one-sided e.g. $H_1: \mu > \mu_0$). 
By the CLT and Slutsky's Theorem, we have that %the test statistic
\vspace{-3pt}
\begin{equation}
    T_\text{one-sample} = \frac{\hat{\mu} - \mu_0}{\SE(\hat{\mu})} = \frac{\sqrt{N}(\hat{\mu} - \mu_0)}{S} \nonumber
\vspace{-3pt}
\end{equation}
follows a standard normal distribution asymptotically. 
We \emph{reject} the null hypothesis $H_0$ at a pre-specified significance level $\alpha$ (usually $\alpha=0.05$ or $0.01$) if $|T| > z_{\nicefrac{\alpha}{2}}$ for the two-sided test, or if $T > z_{\alpha}$ (or $T < -z_{\alpha}$) for the one-sided test. 
Otherwise, we \emph{fail to reject} $H_0$.
This is exactly equivalent to checking if the $(1-\alpha)$ confidence interval excludes $\mu_0$.

Similarly, when comparing two sample means (independent or paired),
testing whether their difference ($\mu_A - \mu_B$ or $\mu_D$) is zero is equivalent to checking if zero lies within the CI.
This equivalence means that both approaches share the same assumptions and limitations, particularly their reliance on asymptotic normality and independent sampling, making them equally unreliable in the small data regimes.

\textbf{Special case: Bernoulli data~~} When $X_i \iid \bernoulli(\theta)$, the variance of the distribution is completely determined by the mean: $\text{Var}(X_i) = \theta (1-\theta)$. Using its sample estimate $S^2 = \hat{\theta}(1-\hat{\theta})$, the standard error simplifies to    
\vspace{-3pt}
\begin{equation}
  \label{eq:bernoulli_se}
    \SE(\hat{\theta}) = \sqrt{\frac{ \hat{\theta}(1-\hat{\theta})}{N}}, \; \text{ where } \hat{\theta} = \frac{1}{N}\sum_{i=1}^N X_i.
\vspace{-3pt}
\end{equation}
This gives us the well-known formula for the confidence interval of a proportion, which has been recommended by and used in several recent  works for LLM evals \citep[e.g.][]{madaan2024quantifying,miller2024adding,dubey2024llama3}.

% \vspace{-10pt}
\section{Failures of the CLT for LLM Evals}
% \vspace{-5pt}
We show that CLT-based CIs break down in many settings in the context of LLM evals with %fewer than a few hundred datapoints. 
small sample sizes.
Each subsection  includes an experiment presenting a different failure mode, following the evaluation protocol outlined~below. % using the following evaluation protocol:

% \dri{BLURB about how computations are done and refer to Appendix for detailsBLURB about how computations are done and refer to Appendix for detailsBLURB about how computations are done and refer to Appendix for details BLURB about how computationsBLURB about how computationsBLURB about how computations}

\textbf{Experimental setup~~}
In each experiment, we sample 100 values of the underlying model performance parameter, $\theta$, from a specified prior distribution. 
% (usually $\betadist(1,1) = \text{Uniform}[0,1]$)
For each $\theta$, we generate 200 independent datasets of sizes $N=3, 10, 30 \text{ and } 100$, giving us a total of 80,000 LLM eval datasets. 
This large number of datasets ensures that our results are robust to the randomness in the data generation procedure. We additionally repeated experiments five times with different random seeds (affecting both the sampled parameters and datasets), resulting in % running all experiments five times with different random seeds leads to 
standard errors in coverage on the order of $10^{-3}$ or lower (see \autoref{app:error_bars_on_error_bars}).
For every dataset, we construct a range of \emph{interval methods} for 100 different nominal coverage levels, $1-\alpha$, ranging from 0.8 to 0.995.

\begin{figure*}[t]
    \centering
    \includegraphics[width=0.87\linewidth]{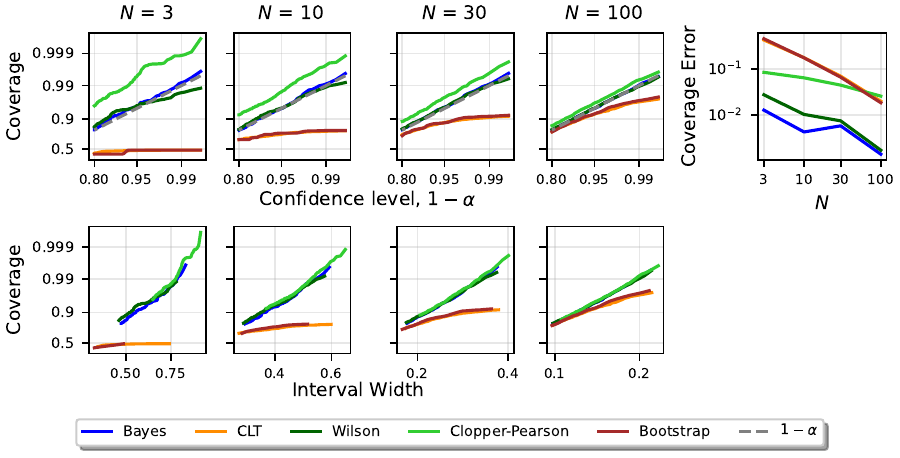}
    \vspace{-10pt}
    \caption{
        \textbf{IID question setting.} Coverage vs. confidence level (\textbf{top}) and vs. interval width (\textbf{bottom}) for various interval-calculation methods on the value of $\theta$. 
        While all methods approach the ideal $1-\alpha$ coverage line for large $N$, only the Bayesian credible interval and Wilson confidence intervals achieve this for small $N$. 
        % Results are averaged over 100 values of $\theta \sim \betadist(1,1)$, each with 200 repeated experiments with randomly generated datasets.
    } 
    \label{fig:iid_intervals}
    \vspace{-11pt}
\end{figure*}

\textbf{Interval methods~~} 
In all experiments, we use at least three methods to construct intervals: 
\begin{itemize}
\vspace{-11pt}
\setlength{\itemsep}{2pt}
\setlength{\parskip}{0pt}
\setlength{\topsep}{0pt}
    \item \textbf{CLT-based} confidence intervals, as described in \autoref{sec:clt_ci}.
    \item \textbf{Bootstrap} confidence intervals which are obtained by resampling the  original data $K$ times (with replacement) to generate a distribution of the estimator $\{\hat{\theta}^{(k)}\}_{k=1}^K$ and taking the empirical $\alpha/2$ and $1-\alpha/2$ quantiles to form the interval. 
    Since bootstrap performance depends greatly on $K$ \citep{davidson2000bootstrap}, we use a large $K=10,000$ throughout.
    \item \textbf{Bayesian} credible intervals derived from the posterior of $\theta$, computed either exactly for conjugate models or approximated using importance sampling. 
    Credible intervals are not unique; 
    in our experiments, we report quantile-based intervals (QBI), either analytic or empirical.  %, computed analytically where possible or empirically otherwise. 
    Highest posterior density intervals (HDI) is a common alternative which we ablate in \autoref{app:f_scores_ablations}. 
    % We show how to do both
    % ; can do HDI or quantiles.
\end{itemize}

\textbf{Evaluation metrics~~}
Frequentist confidence intervals and Bayesian credible intervals are fundamentally different in their interpretation. 
A 95\% confidence interval means that under repeated sampling (as we do here), 95\% of the computed intervals will contain the true $\theta$. 
A 95\% credible interval has a 95\% (posterior) probability of containing $\theta$.
Indeed, confidence intervals are commonly misinterpreted as credible intervals \citep{hubbard2011widespread,greenland2016statistical}.

Despite these differences in interpretation, we evaluate both methods on the same frequentist criterion---\emph{coverage probability}, which we estimate as the empirical proportion of intervals that contain the true $\theta$. 
To highlight differences at the high confidence levels (i.e., 95\% and above), which are most relevant in practice, we use logit-logit axes when plotting empirical versus nominal coverage.
We also report mean absolute distance from nominal coverage, averaged across $\alpha$ and experiment repeats, which we refer to as `coverage error'; this is shown on log-log axes against sample size $N$.
We additionally record the average interval width, which we report in \autoref{app:interval_width}, along with ablations on the choice of the data generating prior in \autoref{app:ablations}.

\begin{figure*}[t]
    \centering
    \includegraphics[width=0.95\linewidth]{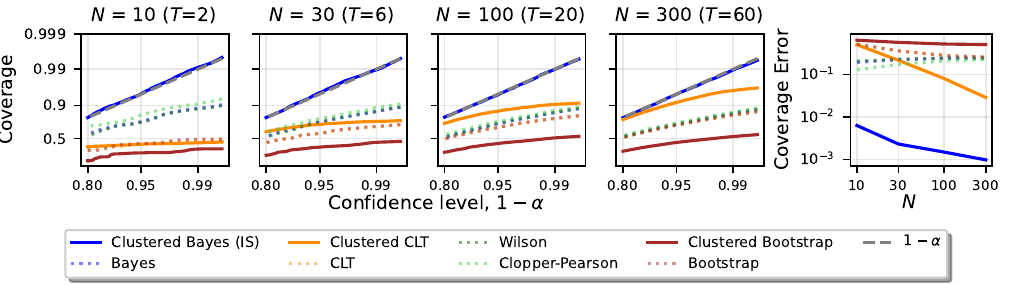}
    \vspace{-4pt}
    \caption{\textbf{Clustered questions setting}. Coverage vs. confidence level for various interval-calculation methods on the value of $\theta$. %(Coverage vs. interval-width plots are given in \autoref{fig:clustered_intervals_with_width}, with further results provided in \autoref{app:clustered_mismatch}.)
    See \autoref{app:interval_width} for interval widths.
    Importantly, note that in a small-data regime, neither simple CLT nor clustered CLT intervals produce correct coverage.
    Methods ignoring the clustered structure of the data %---assuming instead IID questions as per \autoref{sec:failure_simple_ci}---
    are shown as dotted lines.
    %Results are averaged over 100 values of $\theta \sim \betadist(1,1)$, each with 200 repeated experiments with randomly generated datasets.
    }
    \label{fig:clustered_intervals}
    \vspace{-8pt}
\end{figure*}

% \vspace{-10pt}
\subsection{Failure of CLT-Based Confidence Intervals in IID Questions Setting} \label{sec:failure_simple_ci}
% \vspace{-5pt}
% We describe the procedure for obtaining CLT-based confidence intervals in general in \autoref{sec:clt}.
% In the specific setting of LLM evals, the data are Bernoulli, i.e.\ whether or not the LLM correctly answered a particular question,
% where $\theta$ is the underlying probability of getting a question correct, and $i$ ranges from $1$ to $N$.
It is common to assume that benchmarks contain IID questions, so that each eval outcome is a Bernoulli trial with some underlying probability of success $\theta$.
Under this assumption, we compute the standard  standard error of the empirical mean $\hat{\theta}$ using  \autoref{eq:bernoulli_se}.
%\begin{align}
%  \hat{\theta} &= \tfrac{1}{N} \tsum_{i=1}^N y_i\\
%  SE(\^2 &=  \hat{\theta} \b{1-\hat{\theta}}
%\end{align}
However, this expression raises an immediate problem:
as $\hat{\theta}$ approaches 0 or 1, the confidence interval shrinks towards 0, incorrectly suggesting certainty.
In small-data settings, it is perfectly possible that the model gets all questions right ($\hat{\theta} = 1$) or wrong ($\hat{\theta}=0$).
In either case, the CLT-based interval (\autoref{eq:base_CI}) vanishes, since $\SE(\hat{\theta})=0$, which is clearly not right.
% This situation does occur (e.g. with o1's performance on reasoning tasks in LiveBench, \cite{white2024livebench} or Claude's result on \citet{langchain2024benchmarking} Typewriter tool-use benchmark), though it is rare. 
Additionally, when $\hat{\theta}$ or $1-\hat{\theta}$ is less than $1/(1+ \nicefrac{N}{z^2_\alpha})$, the boundaries of the interval would fall outside the valid $[0,1]$ range.
As \autoref{fig:real_data_langchain} shows, both of these problems do occur in practice. % in , which shows error bars calculated with the CLT (\autoref{eq:base_CI}) and alternative methods on the \citet{langchain2024benchmarking} Typewriter tool-use benchmark.

While these issues might occur in practice only rarely, they do highlight that the assumptions underlying the asymptotic, CLT-based approaches may not be suitable for LLM evals, at least in smaller data regimes. In that case, we  expect to see broader failures of CLT-based confidence intervals. %more broadly.
% \begin{figure}[th]
%     % \centering
%     \includegraphics[width=0.95\columnwidth]{fig/real_data_langchain_subset.pdf}
%     \caption{
%         \textbf{Error bars on Langchain Tool-use Benchmark}. The benchmark consists of $N=20$ questions and we show 95\% confidence/credible intervals for the model accuracy, with the empirical mean shown in black and Bayesian posterior mean in blue.
%         The CLT-based approach produces invalid CIs that extend beyond  $[0,1]$ or collapse to zero-width.
%         % Here we report error bars on only 6 models, 
%         % See \autoref{app:real_data} for more results. % with all 9 LLMs.
%     } \label{fig:real_data_langchain}
% \vspace{-10pt}
% \end{figure}

% To test whether we do expect to see such issues in practice, remember that the point of a confidence interval with a particular confidence level, $1-\alpha$, is that, when you sample datasets many times, the true parameter should lie within the confidence interval $1-\alpha$ of the time---i.e. you get the correct \emph{coverage}.
% If we have a toy setting where we can generate data, then we can actually evaluate this coverage property.
To empirically test this hypothesis, we %use a toy setting in which we can directly 
evaluate the coverage of different types of intervals by generating data from
\vspace{-4pt}
\begin{align}
  \theta \sim \betadist(1, 1) \quad %\\
  y_i \sim \bernoulli(\theta) \; \text{for } i=1,\dots N. \label{eq:indep_beta_binom}
\vspace{-4pt}
\end{align}
This simulation framework  mimics the common LLM eval scenario described at the start of this subsection, with true model accuracy  $\theta$ uniformly distributed between $0$ and $1$. %, and for each question, it either answers correctly or incorrectly with probability $\theta$. %the actual outcome is Bernoulli distributed with that probability.
% For every $\theta$ we sample $N$ outcomes, compute frequentist confidence or Bayesian credible intervals and check whether the true $\theta$ lies in the interval. 
% Repeating this $K=10,000$ times allows us to estimate the empirical coverage of these intervals. %and ask what proportion of the time the true $\theta$ lies in the interval.

% Along with CLT-based intervals, 
In addition to the three primary interval methods, here we consider two additional frequentist approaches designed specifically for Bernoulli trials: the approximate Wilson score interval \citep[WS,][]{wilson1927probable} and the exact Clopper-Pearson interval \citep[CP,][]{clopper1934use}, which we describe in \autoref{app:freq_intervals_description} for completeness.
% We also generate confidence intervals via bootstrapping, i.e. subsampling the data with repeats to obtain $K$ synthetic datasets of length $N$ and constructing the interval via the $100(\alpha/2)$ and $100(1-\alpha/2)$ percentiles of the collection of estimates $\{\hat{\theta}^{(k)}\}_{k=1}^K$.
% The quality of bootstrapping greatly depends on the value chosen for $K$ \citep{davidson2000bootstrap}---in order to ensure a fair comparison we use $K=10,000$ throughout.
% Finally, we also use a Bayesian method, in which we report the credible interval
% \footnote{\sam{TODO: briefly explain difference between confidence and credible intervals}} 
% of the posterior distribution of $\theta$.

The results are shown in \autoref{fig:iid_intervals}: the top row plots the confidence level against the empirical coverage.
We find that both CLT-based and bootstrap CIs show poor calibration, as the actual coverage is well below the nominal $1-\alpha$ level,  indicated by the gray dashed line.
This is a fairly catastrophic failure: the whole point of a confidence interval, by construction, is to get the right coverage. %, i.e.\ error bars that capture the right answer $1-\alpha$ of the time.
As an example, in the $N{=}100$ column we see that 95\% CLT-based intervals  achieve a coverage of only 92.5\%.
This leads to a significant difference in interval width: the corresponding Gaussian quantiles (used in \autoref{eq:base_CI}) for confidence 0.95 and 0.925 are $z_{0.025} \approx 1.96$ and $z_{0.0125} \approx 1.78$.
Furthermore, looking at the width of these intervals (bottom row in \autoref{fig:iid_intervals}), we find that CLT-based or bootstrap CIs are inefficient---they are wider than needed for any given coverage.

In contrast, we find that  simple Bayesian methods based on a Beta-Bernoulli distribution, and the frequentist WS interval perform well.
While approximate, WS demonstrates favourable coverage and length properties and is generally preferred to the exact CP, which is overly conservative (too wide) in practice \citep{agresti1998approximate,newcombe2011defence}.
% There is an interesting connection between the exact CP intervals and the Bayesian credible intervals under the uniform $\betadist(1, 1)$ prior, which helps explain the strong empirical performance of the Bayesian approach.
Interestingly, the CP interval can be shown to be equivalent to the Bayesian credible interval but with the uniform prior removed \citep{thulin2014cost}.
In other words, the Bayesian approach can be seen as providing a form of shrinkage of the CP interval thus mitigating over-coverage.

We would, therefore, recommend using WS or Bayesian intervals in practice.
Both methods are easy to apply. WS (and CP) are implemented in SciPy. The posterior for Beta-Bernoulli is available in closed form:
\vspace{-5pt}
\begin{align}
\hspace{-3pt}
  \prob\bc{\theta}{y_{1:N}} &= \betadist\!\b{1+\tsum_{i=1}^N y_i, 1 + \tsum_{i=1}^N (1-y_i)}\!, \hspace{-2pt}\label{eq:beta_binomial_posterior}
\vspace{-10pt}
\end{align}
% Then, you obtain the confidence intervals by taking appropriate quantiles, e.g.\ the range from the $\alpha/2$th quantile to the $1-\alpha/2$th quantile.
so we can use the quantiles of the Beta distribution:
\vspace{-5pt}

\begin{pbox}[label={ex:simple}]{Analysis for a single model}
\begin{minted}{python}
from scipy.stats import binomtest, beta

# y is a length N binary "eval" vector
S, N = y.sum(), len(y) # total successes & questions
result = binomtest(k=S, n=N)

# 95% Wilson score and Clopper-Pearson intervals
wilson_ci = result.proportion_ci("wilson", 0.95)
cp_ci = result.proportion_ci("exact", 0.95)

# Bayesian Credible interval
posterior = beta(1+S, 1+(N-S))
bayes_ci = posterior.interval(confidence=0.95)
\end{minted}
\end{pbox}
% \vspace{-18pt}

%We can also consider a kind of ``data-conditioned confidence interval'', which is the proportion of the time that the confidence interval contains the right $\theta$, given that there were ... correct answers observed in the data.
%Note that is slightly different from the usual frequentist criterion, which specifies that the confidence interval contains the right answer, when we average over all possible datasets.
%You can evaluate this by rejection sampling (i.e.\ you sample a whole bunch of datasets, but take only those datasets where the model got 5 answers right).
%[This is nice way of smuggling in a Bayesian posterior :-)]

% \begin{figure*}[t]
%     \centering
%     \includegraphics[width=0.95\linewidth]{fig/exp4-2_alpha_only.pdf}
%     \caption{\textbf{Clustered questions setting}. Coverage vs. confidence level for various interval-calculation methods on the value of $\theta$. %(Coverage vs. interval-width plots are given in \autoref{fig:clustered_intervals_with_width}, with further results provided in \autoref{app:clustered_mismatch}.)
%     Importantly, note that in a small-data regime, neither simple CLT nor clustered CLT intervals produce correct coverage.
%     Methods ignoring the clustered structure of the data %---assuming instead IID questions as per \autoref{sec:failure_simple_ci}---
%     are shown as dotted lines.
%     %Results are averaged over 100 values of $\theta \sim \betadist(1,1)$, each with 200 repeated experiments with randomly generated datasets.
%     }
%     \label{fig:clustered_intervals}
% \vspace{-10pt}
% \end{figure*}

\begin{figure*}[!t]
    \centering
    \includegraphics[width=0.95\linewidth]{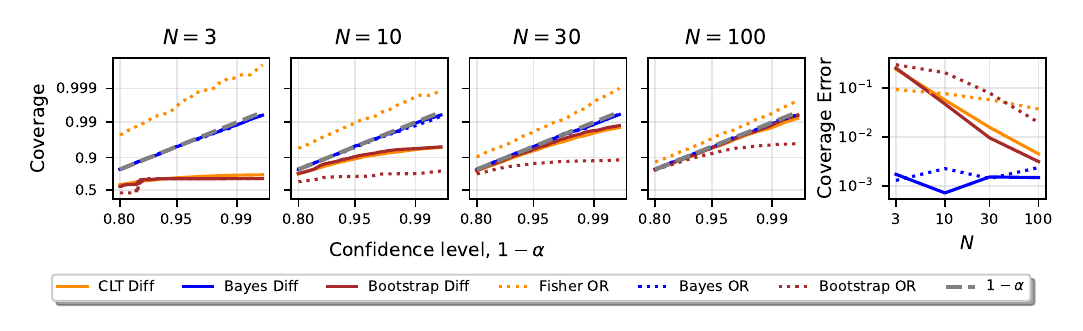}
    \vspace{-10pt}
    \caption{\textbf{Independent model comparison setting}. Coverage vs confidence level for various interval-calculation methods when comparing two independent means $\theta_A$ and $\theta_B$ for both the difference (Diff) and odds ratio (OR) metrics. The diagonal gray dashed line represents the expected coverage, $1-\alpha$. The CLT is not applicable to the OR. 
    % See Appendix~\ref{app:indep_model_comparison} for more results.
    }
    \label{fig:indep_model_comparison}
    \vspace{-5pt}
\end{figure*}

% \subsection{The failure of CLT-based confidence intervals in tasks where questions are clustered}
\subsection{Failure of CLT-Based Confidence Intervals in Clustered Questions Setting}
\label{sec:clt-clustered}
% \vspace{-5pt}

The previous discussion was based on the simplest setting: one LLM and IID questions.
\citet{miller2024adding} emphasizes that the CLT-based approach is flexible enough to apply to more complex settings, %so we look at these in turn here.
%
%In particular, \citet{miller2024adding} considers settings in which 
such as when questions are clustered.
Examples include reading comprehension benchmarks that have multiple questions about a single passage of text \citep[e.g.][]{dua2019drop,choi2018quac,lai2017race,rajpurkar2018know,shi2022language}.
% We would expect some passages of text to be easy for the LLM to understand, implying it should get most questions correct, while the LLM might find some other passages of text harder, implying that it would get most of the questions wrong.
We would expect that LLMs might find some passages of text easier to understand than others, which would result in varying performance---better accuracy on simpler passages and lower accuracy on more complex passages.
Importantly, this introduces non-IID structure in the responses which must be accounted for.
To address this, \citet[][Sec. 2.2]{miller2024adding} suggests using clustered standard errors \citep{abadie2023should}. %, a post-hoc adjustment to the standard error.
This is a post-hoc adjustment to account for the correlation among the $T$ clusters with $N_t$ questions each and $\sum^T_{t=1} N_t = N$:
% In particular, for $T$ clusters each containing $N_t$ questions (with $\sum_{t=1}^T N_t = N$), they suggest
\vspace{-3pt}
\begin{align} %\label{eq:clt_clustered}
    \SE_\text{clust.} = \sqrt{\SE_\text{CLT}^2 + \frac{1}{N^2}\sum_{t=1}^T \sum_{i=1}^{N_t} \sum_{j \neq i} (y_{i,t} - \bar{y})(y_{j,t} - \bar{y})}.\nonumber
\vspace{-5pt}
\end{align}
Here $y_{i,t} \in \{0,1\}$ is the success on question $i$ of task $t$, and $\bar{y} = \frac{1}{N}\sum_{t=1}^T\sum_{i=1}^{N_t} y_{i,t}$.
To assess the effectiveness of this approach, we use data from the following generative~model:
% \begin{subequations}
\vspace{-10pt}
\begin{align}
\begin{split}
  d &\sim \gammadist(1, 1), \quad \theta \sim \betadist(1, 1)  
   \\
  \theta_t &\sim \betadist(d \theta, d (1-\theta)), \quad
  y_{i,t} \sim \bernoulli(\theta_t). \label{eq:clsutered_model}
\vspace{-10pt}
\end{split}
\end{align}
% \end{subequations}
Here, $d$ controls the the range of difficulties of the tasks or clusters (i.e.\ the \emph{concentration}), $\theta$ is the global performance of the model, and is the quantity we are trying to infer, while $\theta_t$ is the performance on a given task.  
Note that  $\E\sqb{\theta_t \halfmid \theta} = \theta$, so $\theta$ controls the expected accuracy on any given task.
If $d$ is large then $\theta_t$ is always close to $\theta$, which indicates lower correlations between questions within a task.
In contrast, if $d$ is small, then $\theta_t$ is further from $\theta$, implying larger correlations between questions in a task/cluster.

% To perform inference in this model, first let $Y_t$ be the total number of correct answers in task $t$. 
% Integrating out $\theta_t$ gives
% % To perform Bayesian inference in \autoref{fig:clustered_intervals}, we integrate out $\theta_t$,
% % \begin{subequations}
% % \begin{align}
% %   \theta &\sim \betadist(1, 1)\\
% %   d &\sim \text{Gamma}(1, 1)\\
% %   Y_t &\sim \text{BetaBinomial}(N_t, d \theta, d (1-\theta))
% % \end{align}
% % \end{subequations}
% % \vspace{-8pt}
% \begin{align}
%   Y_t &\sim \text{BetaBinomial}(N_t, d \theta, d (1-\theta)).  \nonumber %\label{eq:clustered_IS}.
%   % \vspace{-6pt}
% \end{align}
% % Now there are only two latent variables, $\theta$ and $d$, meaning that the inference problem is fairly straightforward, and amenable to many different approaches.
To perform inference in this model, we can integrate out $\theta_t$ so that the total number of correct answers in task $t$ is given by $Y_t \sim \betabin(N_t, d \theta, d (1-\theta))$.
For simplicity, we use importance sampling (IS) with the prior as proposal: \(\theta \sim \beta(1, 1)\) and \(d \sim \gammadist(1,1)\).
% \begin{subequations}
% \begin{align}
%   \Q\b{\theta} &= \betadist(\theta; 1, 1)\\
%   \Q\b{d} &= \text{Gamma}(d; 1, 1).
% \end{align}
% \end{subequations}
Then the importance weights $\{w^{(k)}\}_{k=1}^K$ are  given by the Beta-Binomial likelihood of the data $Y_t$ under each $(\theta^{(k)}, d^{(k)})$: %, computed using the Beta-Binomial in \autoref{eq:clustered_IS}
% (of which we draw $K=10,000$) 
% using the pdf of the beta-binomial in \autoref{eq:clustered_IS}, which is available in closed-form.
\vspace{-4pt}
\begin{align}
    w^{(k)} = \tprod_{t=1}^T \betabin(Y_t; N_t, d^{(k)} \theta^{(k)}, d^{(k)}(1-\theta^{(k)})). \nonumber
\vspace{-6pt}
\end{align}
We use these weights to resample (with replacement) our collection of samples $\{\theta^{(k)}\}_{k=1}^K$ and calculate credible intervals using the relevant percentiles in the resulting collection.
Simple Python code for this IS is provided in Appendix \ref{app:clustered_bayes_code}.

% To our knowledge there are no readily available frequentist tests that do not use the CLT or bootstrap available for this setting.  
% We therefore compared Bayesian, CLT and bootstrap methods based on clustered and IID models, along with methods such as Clopper-Pearson, which are based on an IID model (Fig.~\ref{fig:clustered_intervals}).
% We found that only the Bayesian method based on a clustered model gave the right coverage.

\autoref{fig:clustered_intervals} compares the three primary methods under both the IID and clustered question assumption along with the CP and WS intervals (which also assume IID data).
To our knowledge there are no readily available frequentist methods, tailored to such clustered data that can be applied in here. 
Of all interval calculation methods we consider, only the Bayesian method based on a clustered model (\autoref{eq:clsutered_model}) achieves the right coverage across different sample sizes.

% \sam{I think the importance sampling code is too long to put in the main text} \dri{agree}

% \begin{figure*}[t]
%     \centering
%     \includegraphics[width=0.95\linewidth]{fig/indep_model_comparison.pdf}
%     \vspace{-10pt}
%     \caption{\textbf{Independent model comparison setting}. Coverage vs confidence level for various interval-calculation methods when comparing two independent Bernoulli parameters $\theta_A$ and $\theta_B$ for both the difference (Diff) and odds ratio (OR) metrics. The diagonal gray dashed line represents the expected coverage, $1-\alpha$. The CLT is not applicable to the OR. 
%     % See Appendix~\ref{app:indep_model_comparison} for more results.
%     }
%     \label{fig:indep_model_comparison}
% \vspace{-10pt}
% \end{figure*}

\begin{figure*}[!t]
    \centering
    \includegraphics[width=0.95\linewidth]{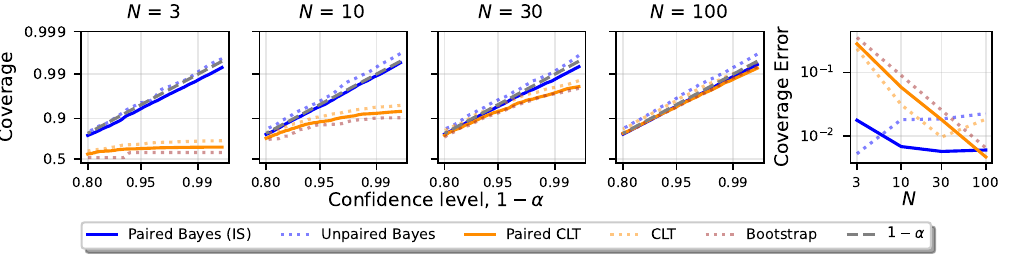}
    \vspace{-10pt}
    \caption{\textbf{Paired model comparison setting}. Coverage vs. confidence level for various interval-calculation methods on the value of $\theta_A - \theta_B$.
    % (Coverage vs. interval-width plots are given in \autoref{fig:paired_intervals_with_width}, with further results provided in \autoref{app:paired_mismatch}.)
    Methods ignoring the paired structure of the data---assuming instead IID questions and answers from model A and from model B, as per \autoref{sec:failure_simple_ci}---are shown as dotted lines.
    % Results are averaged over 100 values of $\theta_A, \theta_B \sim \betadist(1,1)$, each with 200 repeated experiments with randomly generated datasets.
    }
    \label{fig:paired_intervals}
    \vspace{-10pt}
\end{figure*}

% \vspace{-10pt}
\subsection{Failure of CLT-Based Confidence Intervals in Independent Model Comparison Setting} \label{sec:indep_model_comparison}
% \vspace{-5pt}
% \dri{this is sectio 4.1 in Miller}

We often wish not only to construct a confidence interval for the performance of a single model, but also to compare two models. 
Consider the setup from \autoref{sec:failure_simple_ci} but now we have two language models, $A$ and $B$. 
Let $\theta_A$ and $\theta_B$ be the true probabilities of success of models $A$ and $B$, each of which is independently generating Bernoulli outcomes as in the model defined in \autoref{eq:indep_beta_binom}. 
This setup is applicable, for example, when we only have access to the empirical accuracies, $\hat{\theta}_A$ and $\hat{\theta}_B$, and not the per-question binary data, itself ($y_{A; i}$ and $y_{B; i}$), or when models are evaluated on different sets of questions.
In these cases, there are two main approaches for comparing $\theta_A$ and $\theta_B$: 
\begin{itemize}
% \vspace{-8pt}
\setlength{\itemsep}{3pt}
\setlength{\parskip}{0pt}
\setlength{\topsep}{0pt}
    \item looking at their \emph{difference}, $\operatorname{Diff} = \theta_A - \theta_B$, and checking if 0 lies within its $(1-\alpha)$ confidence interval.  %it is statistically equal to 0 (i.e. is contained in the CI).
    \item looking at their \emph{odds ratio}, $\operatorname{OR} = \frac{\theta_A / (1 - \theta_A)}{\theta_B / (1 - \theta_B)}$, and checking if 1 lies within its $(1-\alpha)$ confidence interval.
% \begin{equation}
    % \text{OR}_{\theta_A, \theta_B} = \frac{\theta_A / (1 - \theta_A)}{\theta_B / (1 - \theta_B)}.
% \end{equation}
%and checking if 1 lies within its CI% it is statistically equal to 1.
\end{itemize}
\vspace{-7pt}
A confidence interval including the respective null value (0 for the difference, 1 for the odds ratio) suggests that $\theta_A$ and $\theta_B$ are statistically indistinguishable at level $\alpha$.

With the CLT, we can only construct a confidence interval on the difference in performances, since the odds ratio is a non-linear transformation of the parameters. 
Whilst the CLT guarantees asymptotic normality for the difference of proportions, it does not extend to their ratio or odds ratio (more on this in \autoref{sec:metrics_not_averages}).
% The construction of this CI was described in \autoref{eq:2indep_sample_clt_ci}.
As shown by the solid orange curves in \autoref{fig:indep_model_comparison}, the CLT-based interval from \autoref{eq:2indep_sample_clt_ci} has coverage far below the target level when $N$ is small. %\footnote{
Although better frequentist methods exist, e.g. the hybrid score interval introduced in \citet{newcombe1998interval}, they are harder to implement and are not available in standard libraries. %}

For odds ratio analysis, the standard frequentist method to obtaining confidence intervals involves inverting the Fisher’s exact test \citep[FET,][]{fisher1922interpretation}, which \emph{guarantees} coverage of at least $1-\alpha$ at any dataset size and any pair of parameters. 
However, similarly to the CP exact interval for a single proportion, FET tends to be overly conservative, especially for small $N$, as shown by the dotted orange curves in \autoref{fig:indep_model_comparison}.

The Bayesian approach is able to give us credible intervals for both the difference and the odds ratio.
To construct these, we draw samples from the exact posterior (\autoref{eq:beta_binomial_posterior}), compute the metric of interest and take the empirical quantiles. %or highest density ratio (HDI).
% As shown by the blue curves in \autoref{fig:indep_model_comparison}, 
As \autoref{fig:indep_model_comparison} shows, the Bayesian intervals achieves excellent coverage across all sample sizes for both metrics.

% \citet{altham1969exact} showed an interesting connection between the Fisher's exact test and the independent Beta-Binomial model we use here. 
% Specifically, the frequentist test  corresponds to the Bayesian analysis using highly conservative priors, $\theta_A \sim \betadist(1, 0)$ and $\theta_B \sim \betadist(0, 1)$. 
% These improper priors effectively assume the most extreme scenario---perfect performance for model A and worst possible performance for model B, which helps explain the overcoverage.
\citet{altham1969exact} showed that Fisher's exact test corresponds to our Bayesian analysis if we use highly conservative priors, $\theta_A \sim \betadist(1, 0)$ and $\theta_B \sim \betadist(0, 1)$. 
These improper priors effectively assume the most extreme scenario---perfect performance for model A and worst possible performance for model B, which helps explain the over-coverage.

% [TODO] here we also care about detecting a difference. Figure [TODO] shows: given $\theta_A \neq \theta_B$, are we failing to reject the Null? Plot $|\theta_A - \theta_B|$ vs proportion of time $0$ (or $1$) is in the CI.

% \paragraph{More advanced frequentist methods} 
% There are more advanced frequentist, methods such as Newcombe’s hybrid score interval \citep{newcombe1998interval} that substantially improve on the performance of the CLT-based interval, but are not available in common libraries.
% Additionally, the Boschloo test \citep{boschloo1970raised} offers an improvement over Fisher's exact test, making it less conservative, and is implemented in common libraries (like SciPy).
% However, inverting the Boschloo test to construct confidence intervals difficult and rarely implemented. 
% For completeness purposes, we show it does work better than fisher by plotting the p-values... [probably in Appendix] 

% \paragraph{Recommendation}   
With small data, we recommend Bayesian intervals for the difference or odds ratio. 
% As the next code snippet shows, the implementation is again simple.
It can be implemented as follows:
\vspace{-5pt}

\begin{pbox}[label={ex:bayes_comparison}]{Bayesian analysis: Model comparison}
% \begin{listing}[!ht]
\begin{minted}[fontsize=\scriptsize]{python}
# y_A and y_B are vectors of evals for two models
import numpy as np

S_A, S_B = y_A.sum(), y_B.sum()
# draw posterior samples (ps)
ps_A = beta(1 + S_A, 1 + (N - S_A), size=2000)
ps_B = beta(1 + S_B, 1 + (N - S_B), size=2000)

# posterior difference and 95% QBI
ps_diff = ps_A - ps_B  
bayes_diff = np.percentile(ps_diff, [2.5, 97.5]) 

# posterior odds ratio and 95% QBI
ps_or = (ps_A / (1 - ps_A)) / (ps_B / (1 - ps_B))
bayes_or = np.percentile(ps_or, [2.5, 97.5]) 
\end{minted}
% \end{listing}
\end{pbox}
% #  odds_ratio(...).confidence_interval(0.95)

\vspace{-3pt}
\begin{remark}[\textbf{Bayesian model comparison}]\label{remark:bayes_model_comparison}
    A key benefit of the  Bayesian approach to model comparison is that we can easily compute probabilities that one model outperforms another.
    Given posterior samples, $\theta_m^{(k)} \sim p(\theta_m \halfmid y_{m, 1:N})$, for models $m \in \{A, B\}$, we can calculate:
\vspace{-9pt}
\begin{equation}
    \mathbb{P}(\theta_A > \theta_B \halfmid y_{A;1:N}, y_{B;1:N}) = \frac{1}{K} \sum_{k=1}^K \mathbbm{1}[\theta_A^{(k)} > \theta_B^{(k)}]. \nonumber
% \vspace{-15pt}
\end{equation}
We cannot make such probabilistic statements in frequentist inference since the parameters are treated as fixed (but unknown) constants and probabilities  only refer to hypothetical repetitions of the data.
\end{remark}

\begin{figure*}[!t]
    \vspace{-8pt}
    \centering
    \includegraphics[width=0.95\linewidth]{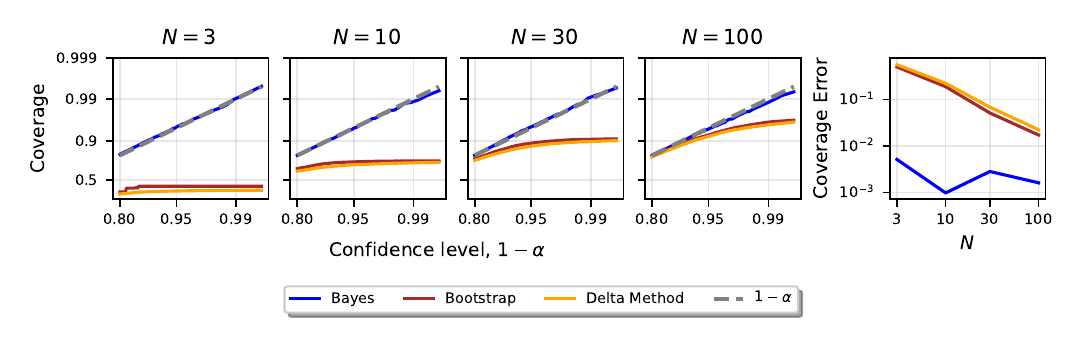}
    \vspace{-12pt}
    \caption{\textbf{$F_1$-score error bars}. 
    Coverage vs. confidence level for Bayesian and bootstrap intervals. The CLT is not directly applicable due to the non-linearity of the $F_1$-score, so we use the the delta method to construct a CLT-based CI.
    }
    \label{fig:f_scores}
    \vspace{-10pt}
\end{figure*}

% \vspace{-10pt}
\subsection{Failure of CLT-Based Confidence Intervals in Paired Model Comparison Settings}\label{sec:paired}
It is important to also consider model comparison in the paired setting, where two models have been evaluated on the \textit{same} set of questions.
%Naively, 
A natural approach is to construct a CLT-based interval on $\theta_A - \theta_B$ using \autoref{eq:paired_clt_ci} which directly takes into account the paired structure of the data. 
%However, in practice, the CLT is inappropriate here as it assumes IID data, whereas the answers of two models to the same question are not independent. 
Paired intervals are advantageous because evaluating both models on the same questions lets common  question-specific effects cancel out, which reduces variance. 
This leads to a more precise comparison if the outcomes are positively correlated, though negative correlation might increase variance.
To simulate paired evals data, we first sample probabilities of success for each model, alongside a correlation %\footnote{Note that this formulation encourages positive correlations, which are more likely in practice. See \autoref{app:paired_simulation} for details.} 
$\rho = 2\hat{\rho}-1$:
\begin{align}
    \theta_A, \theta_B \sim \betadist(1,1), \quad
    \hat{\rho} \sim \betadist(4,2). \nonumber
\end{align}
This formulation encourages positive correlations, which is expected in the context of LM evals: when models are evaluated on the same questions, factors such as question difficulty tend to affect both models similarly, leading to positively correlated performance (more details in \autoref{app:paired_simulation}).

Next, we sample $N$ points from a bivariate Gaussian
\begin{align}
    (a_1,b_1), \ldots, (a_N, b_N) \iid \mathcal{N}\left(\begin{pmatrix} \Phi^{-1}(\theta_A) \\ \Phi^{-1}(\theta_B)\end{pmatrix}, \begin{pmatrix}1 & \rho \\ \rho & 1\end{pmatrix} \right), \nonumber
\end{align}
where $\Phi(\cdot)$ is the standard univariate Gaussian CDF.
This parameterisation of the Gaussian covariance ensures that each 2D point has unit variances and correlation $\rho$.
Meanwhile, the choice of the Gaussian's mean ensures that if we obtain binary eval outcomes for model A and model B by considering the sign of $a_i$ and $b_i$ respectively:
\vspace{-5pt}
\begin{align}
    y_{A;i} = \mathbbm{1}[a_i > 0], \quad
    y_{B;i} = \mathbbm{1}[b_i > 0], \nonumber
\vspace{-5pt}
\end{align}
and the marginal probabilities of success for both models are as desired, that is, $\mathbb{P}(y_{A;i}) = \theta_A$ and $\mathbb{P}(y_{B;i}) = \theta_B$.
% Further details are provided in \autoref{app:paired_simulation}.

Much like in the clustered setting (\autoref{sec:clt-clustered}), we perform Bayesian inference on the posterior distribution of $\theta_A - \theta_B$ using importance sampling, drawing $K\!=\!10,000$ samples from the prior as the proposal distribution.
Details and code for this can be found in Appendix \ref{app:paired_bayes_code}.
We also present an `unpaired Bayes' method in which we construct credible intervals on $\theta_A - \theta_B$ by sampling from the posteriors obtained for each of $\theta_A$ and $\theta_B$ separately as in \autoref{sec:indep_model_comparison}.%using the Beta-Binomial model in \autoref{sec:failure_simple_ci}.

We see in \autoref{fig:paired_intervals} that all non-Bayesian methods severely underperform when it comes to achieving nominal coverage for small $N$.
Moreover, the advantages  of Bayesian inference for enabling direct probabilistic comparison, as discussed in Remark~\ref{remark:bayes_model_comparison}, apply equally well in this setting.
We would recommend using the paired Bayes method as it can account for correlations and thus produce narrower intervals. The unpaired Bayes is also a reasonable and easier to implement alternative  %as opposed to the unpaired version due to its more robust behaviour when dealing with mismatched priors 
(see Appendix \ref{app:paired_mismatch} for ablations).

% \vspace{-10pt}
\subsection{Failure of CLT-Based Confidence Intervals When Metrics Are Not Simple Averages}\label{sec:metrics_not_averages}
% \vspace{-5pt}

Many metrics for LLM evals cannot be represented as simple averages of IID variables, in which case the CLT cannot be used to construct confidence intervals at all.
We already saw this issue with the odds ratio in \autoref{sec:indep_model_comparison}, but the problem extends to many other widely used metrics. % \end{matrix}$
Indeed, some recent work, e.g. the Llama~3 report, acknowledges this limitation and omits reporting confidence intervals for metrics that are not simple averages \citep[see p.29 in][]{dubey2024llama3}.

% For such metrics, Bayesian approaches offer a natural framework for uncertainty quantification. 
For many tasks (e.g. retrieval), it makes sense to not only track whether a model gave a correct or incorrect response, but also whether that response was a true positive (TP), true negative (TN), false positive (FP), or a false negative (FN). 
These counts form a 2$\times$2 contingency table known as a confusion matrix.  
The outcome $y_i$ can therefore be viewed as a draw from a Categorical distribution with some parameter $\bm{\theta} \coloneqq (\theta_\text{TP}, \theta_\text{FP}, \theta_\text{FN}, \theta_\text{TN})$, and the total counts in each category $N_\text{conf} \coloneqq (N_\text{TP}, N_\text{FP}, N_\text{FN}, N_\text{TN})$ is a draw from a $\multinomial(N, \bm{\theta})$.
To simulate an evaluation dataset we sample ground truth parameters from a uniform Dirichlet prior, which is conjugate to the categorical likelihood: % giving us a closed form posterior as follows: %place an uninformative Dirichlet prior on $\bm{\theta}$. The model we generate data from is:
\vspace{-3pt}
\begin{align}
\begin{split}
    \bm{\theta}  &\sim \dirichlet(1, 1, 1, 1),  \\ 
    y_i &\sim \categorical(\bm{\theta}),  \\
    \bm{\theta} \halfmid y_{1:N} &\sim \dirichlet(1 + N_\text{conf}).  
\end{split} \label{eq:dirichlet_multinomial}
\vspace{-3pt}
\end{align}
% Here, each $y_i$ represents a single observation taking one of the four possible confusion matrix outcomes. 
% The conjugate Dirichlet posterior is obtained by adding the observed counts to the prior parameters
% \begin{equation}
% \bm{\theta} \halfmid y_{1:N} \sim \dirichlet(1 + N_\text{TP}, 1 + N_\text{TN}, 1 + N_\text{FP}, 1 + N_\text{FN}). \nonumber
% \end{equation}
Many metrics derived from the confusion matrix, e.g. $F_\beta$-scores, MCC or G-score, are non-linear in $\bm{\theta}$, so the CLT is not applicable~\citep{caelen2017bayesian}.
Under standard regularity conditions, the delta method \cite{oehlert1992note} can be used to in conjunction with the CLT to approximate the sampling distribution of smooth non-linear functions of estimators via a first-order Taylor expansion (see \autoref{app:freq_intervals_description} for details).

As an example, consider the $F_1$ score, which is the harmonic mean of precision and recall:
% \begin{equation}
%     F_1 = 2\,\frac{\precision\cdot\recall}{\precision+\recall}, \nonumber % = \frac{}{},
% \end{equation}
%where 
 % are defined in terms of the confusion matrix counts. 
\vspace{-2pt}
\begin{equation}
    F_1 = 2\,\frac{\precision\cdot\recall}{\precision+\recall}, % = \frac{}{},
    \vspace{-2pt}
\end{equation}
where $\precision\!=\! \frac{N_\text{TP}}{N_\text{TP} + N_\text{FP}}$ and $\recall = \frac{N_\text{TP}}{N_\text{TP} + N_\text{FN}}$.

% To the best of our knowledge,  no readily available alternative frequentist methods exist for constructing confidence intervals for $F_1$. 
Our  empirical evaluation is therefore focused only on comparing Bayesian credible intervals based on the model from \autoref{eq:dirichlet_multinomial} against the bootstrap, with results presented in 
\autoref{fig:f_scores}. 
The Bayesian intervals closely track the nominal coverage, while the bootstrap ones systematically under-cover. 
We therefore recommend using Bayesian intervals in practice. 
The next code snippet demonstrates how to implement this approach.
To show robustness to the choice of interval, we also include highest density intervals (HDIs), with results shown in \autoref{fig:f_scores_hdi} in the Appendix.
% in including an option for highest density intervals (HDIs), in addition to the quantile-based ones we have used throughout.  
% We provide additional results for HDIs in \autoref{fig:f_scores_hdi} in the Appendix to show robustness to the choice of interval.

% \vspace{-5pt}

\begin{table*}[!th]
    \centering
    \caption{\textbf{Summary of interval methods and their key properties}. 
    \emph{Coverage} describes whether a method is able to provide at least the desired nominal coverage in small-sample settings.
    \emph{Efficiency} describes how tight and precise the resulting  intervals are given the nominal coverage (e.g., CLT-based intervals can be invalid or too wide).
    The \emph{computational cost} of all methods is negligible compared to the cost of running the LLM evals, though we indicate their relative costs to facilitate comparison across methods.
    % \\ (*) Note: The Delta method works only for \textit{differentiable} functions.
    }
    \vspace{-10pt}
    \begin{tabular}{rcccc}
     & \makecell{Coverage\\ small $N$} & \makecell{Efficiency\\ small $N$} & \makecell{Computational\\ cost} & \makecell{Easy to\\ implement} \\%& \makecell{Handles non-linear \\ functions of params} \\ 
     \toprule
     CLT & \xmark & \xmark & Very low & Yes \\%& \xmark \\ 
     CLT-based variants (e.g. Delta method) & \xmark & \xmark & Very low & Moderate \\%& \cmark\\
     Custom frequentist (e.g. Wilson) & \cmark & \cmark & Very low & Moderate \\%& \xmark \\
     Bootstrap & \xmark & \xmark & Low & Moderate \\%& \cmark \\
     Bayes (conjugate) & \cmark & \cmark & Very low & Yes \\%& \xmark \\
     Bayes (importance sampling) & \cmark & \cmark & Low & Moderate \\%& \cmark \\ 
     \bottomrule
    \end{tabular}
    \label{tab:summary}
\end{table*}

\begin{pbox}[label={ex:f_score}]{Bayesian credible interval for $F_1$ score}
\begin{minted}[fontsize=\scriptsize]{python}
from numpy.random import dirichlet
from arviz import hdi

# confusion_arr is np.array([N_TP,N_FP,N_FN,N_TN])
ps = dirichlet(confusion_arr + 1, 2000) 
f1_samples = calculate_f1(ps) # implements Eq.10

# 95% HDI and QBI
bayes_hdi = hdi(f1_samples, hdi_prob=0.95)
bayes_qbi = np.percentile(f1_samples, [2.5, 97.5]) 
\end{minted}
\end{pbox}
\vspace{-10pt}

% \vspace{-10pt}
\section{Alternative Views}
% \vspace{-5pt}
It may be argued that CLT-based methods are usually sufficient in practice when their assumptions are satisfied.
% due to their simplicity and computational efficiency
% when their---relatively minimal---assumptions are satisfied.
% For instance, \citet{miller2024adding} states that ``the Central Limit Theorem is applicable to any evals having scores with finite variance and a large number of questions.''
% This is clearly true, and indeed for large $N$ our experiments show that the CLT methods proposed in \citet{miller2024adding} work well and represent a marked improvement upon naive CLT error bars.
We do not disagree. 
However, we  argue that it is safer to use the more robust strategies laid out in this paper, which are just as easy to apply (as demonstrated throughout), perform no worse for large $N$ and perform substantially better in the increasingly common small-$N$ setting. 
% In choosing this paper's title position as ``Don't Use the CLT in LLM Evals With Fewer Than a Few Hundred Datapoints", rather than something more specific, such as ``Don't Use the CLT [...] With Fewer Than 100 Datapoints", we hope to encourage researchers to 
This is especially important since knowing whether a certain $N$ is ``large enough'' for the CLT to hold---a vague and unhelpful framing of the problem---would be extremely context-dependent and difficult to determine \textit{a priori}.
A key reason we chose to word the title of this paper as we did is to avoid implying the existence of some hard-valued $N^*$ 
below which the CLT is always invalid and unjustifiable, and
above which it is always sound and reliable.
% at which the CLT switches from being an unsound, underperforming method to a sound and robust one.) 

% \citet{miller2024adding} goes on to say ``and so we regard bootstrapping as unnecessary unless a complicated sampling scheme or estimator is being used.''
% This brings us to a second alternative view which is well worth considering: CLT error bars should be used because of their simplicity and computational efficiency. 
% It is true that the bootstrapping and Bayesian methods are more complicated to implement and computationally expensive than both the naive and extended versions of CLT-based CIs (\sam{TODO: table in appendix with run-times per method?}), however, we would disagree that this difference is large enough to entirely disregard the benefits of Bayes (which we see outperforming bootstrapping), particularly for small $N$.
% Additionally, we argue that, at least in the common IID question setting with binary outcomes, Wilson score intervals offer a robust and efficient alternative to CLT-based error bars that still lie in the frequentist setting and are extremely easy to implement via well-known packages such as \verb|scipy|.

We argued that Bayesian credible intervals can be useful and flexible alternatives to CLT-based confidence intervals, particularly in settings where other frequentist methods are either too complicated or non-existent. 
Two common criticisms of Bayesian methods are that they are sensitive to the choice of a prior and that they can be computationally expensive.
It is worth exploring these points in some detail with respect to the Bayesian methods discussed above.

Throughout this paper we use broad, non-informative, uniform priors over model performance to ensure that error bars are determined  only by the benchmark results, and not by strong prior beliefs about how we might expect a certain model to perform.
Whilst incorporating such information into an \emph{informative} or \emph{subjective} prior can lead to tighter error bars, it comes with important caveats.
First, achieving optimal coverage and efficiency (i.e. producing the smallest interval width) requires knowing the true underlying prior information \citep{severiniBayesianIntervalEstimates1993}. 
In practice, this information is typically unavailable or unreliable. 
% \footnote{In fact, achieving optimal coverage and efficiency (minimal interval width) \textit{requires} knowing the true underlying prior information \citep{severiniBayesianIntervalEstimates1993}.}, such information might not be available or reliable.
Second, subjective priors are often viewed as controversial: the choice of a specific prior can be arbitrary or unjustified, which in turn will introduce unwanted biases \citep{efronWhyIsntEveryone1986, gelmanObjectionsBayesianStatistics2008}.

Nonetheless, considering the impact of priors is critical in any Bayesian procedure.
In \autoref{app:ablations} we explore the effect of prior mismatch across each of the experimental settings (\S \ref{sec:failure_simple_ci}--\ref{sec:metrics_not_averages}).
In these ablations, we continue to use our non-informative uniform prior for inference, but consider scenarios where we \textit{could} incorporate additional prior information. 
For example, \autoref{fig:iid_intervals_beta_100_20} and  \autoref{fig:iid_intervals_beta_20_100} consider the cases where a model is expected to perform well or badly, respectively. %where we can expect a model to perform generally well or whilst \autoref{fig:iid_intervals_beta_20_100}. 
We find that Bayesian coverage performance generally does not fall below that of CLT-based methods and often still outperforms them, especially for small $N$.

Finally, the added computational cost of Bayesian 
inference is negligible compared to the overall cost of benchmark construction and running the LLM evals themselves (see \autoref{app:compute_time}), making it a worthwhile trade-off for improved accuracy in uncertainty quantification.
In \autoref{tab:summary} we summarise these comparisons in terms of interval quality, computational cost and ease of implementation.

% \vspace{-10pt}
\section{Conclusion}
% \vspace{-5pt}
% We argue against using CLT-based (and bootstrap-based) CIs for LLM eval error bars when the number of datapoints, $N$, is small.
% This is because the assumptions used by the CLT---independence of samples and large $N$---are often not applicable to LLM evals.
% Many eval datasets have highly structured correlations between questions, models have correlation between their answers, and it is increasingly necessary to perform evals on datasets with small $N$.
% Instead, we argue for methods which do not rely on the same set of assumptions: in particular Wilson score confidence intervals and Bayesian credible intervals.
% % We hope that these methods become more common in evaluating LLMs as we've shown them to perform favourably compared to CLT- and bootstrap-based CIs.
% % Although slightly more complicated than naive CLT CIs, 
% These are easily implementable (with example code provided herein) and should become a fixture of LLM evals for their superior uncertainty quantification and---particularly in the case of Bayesian credible intervals---for their flexibility and interpretability.

% \dri{I've crisped up the above + added the point about other metrics and that the bootstrap also does poorly ( we haven't argued against it, but we havent recommended it either). It the below we are at length but have a look and make any changes if you want!}
% \sam{Looks great!}

In this position paper, we argued against using the CLT~to construct confidence intervals for LLM evals because the assumptions---a \emph{large} number of \emph{independent} samples---are rarely satisfied. 
LLM evals often have highly structured correlations among questions, 
correlated model outputs, 
and rely on increasingly smaller, specialized benchmarks.
The CLT also does not apply to common metrics like $F$-scores that are not simple averages of IID variables. 
Of the alternatives that we examined, we found that boostrap intervals also perform poorly, while more appropriate frequentist methods and Bayesian credible intervals are much more reliable.
We provided examples and code demonstrating how easy it is to implement these methods, and we recommend adopting them as standard practice for modern LLM evaluations.

% \vspace{-10pt}
\section*{Acknowledgements}
% \vspace{-5pt}
Sam Bowyer is supported by the UKRI EPSRC via the COMPASS CDT at the University of Bristol (EP\/S023569\/1). 
This work was possible thanks to the computational facilities of the University of Bristol’s Advanced Computing Research Centre---\url{http://www.bris.ac.uk/acrc/}.
We would like to thank Dr. Stewart for funding compute resources used in this project.
DRI would like to thank Gregorio Benincasa for a helpful discussion.
% \clearpage
% \newpage

\bibliographystyle{icml2025}
\bibliography{refs}

%%%%%%%%%%%%%%%%%%%%%%%%%%%%%%%%%%%%%%%%%%%%%%%%%%%%%%%%%%%%%%%%%%%%%%%%%%%%%%%
%%%%%%%%%%%%%%%%%%%%%%%%%%%%%%%%%%%%%%%%%%%%%%%%%%%%%%%%%%%%%%%%%%%%%%%%%%%%%%%
% APPENDIX
%%%%%%%%%%%%%%%%%%%%%%%%%%%%%%%%%%%%%%%%%%%%%%%%%%%%%%%%%%%%%%%%%%%%%%%%%%%%%%%
%%%%%%%%%%%%%%%%%%%%%%%%%%%%%%%%%%%%%%%%%%%%%%%%%%%%%%%%%%%%%%%%%%%%%%%%%%%%%%%
\newpage
\appendix
\onecolumn

\section{Frequentist Confidence Intervals} \label{app:freq_intervals_description}

\subsection{Confidence Intervals Based on the t-Distribution} 

When the data is normally distributed with unknown mean and variance, $X_1, \dots, X_N \iid \gN(\mu, \sigma^2)$, the Student's t-distribution provides an exact \emph{finite} sample solution for confidence intervals (and hypothesis tests):
\begin{equation}
    \CI_{1-\alpha}(\mu) = \hat{\mu} \pm t_{\nicefrac{\alpha}{2}, \nu}\,\SE(\hat{\mu}), \label{eq:t_based_ci}
\end{equation}
where $t_{\nicefrac{\alpha}{2}, \nu}$ is the $\nicefrac{\alpha}{2}$-th quantile of the Student's t-distribution with $\nu=N-1$ degrees of freedom. 
When $N \approx 30$ (or above), the t-distribution is close to the standard normal (e.g. $t_{0.025, 29} = 2.045$ vs  $z_{0.025} = 1.960$).

Confidence intervals based on \autoref{eq:t_based_ci} are often used even when the data is not normally distributed.
In this case, the following assumptions are required for exactness:
\begin{itemize}
    \item The sample mean, $\hat \mu$, is approximately normally distributed (which by the CLT holds for large enough sample sizes).
    \item The quantity $\frac{S^2 (N-1)}{\sigma^2}$ follows a Chi-Squared distribution with $N-1$ degrees of freedom, $\chi^2(N-1)$, that is \textbf{independent} of the sample mean, $\hat \mu$.
\end{itemize}

\begin{figure}[htbp]
    \centering
    \begin{subfigure}[b]{0.99\linewidth}
        \centering
        \includegraphics[width=\linewidth]{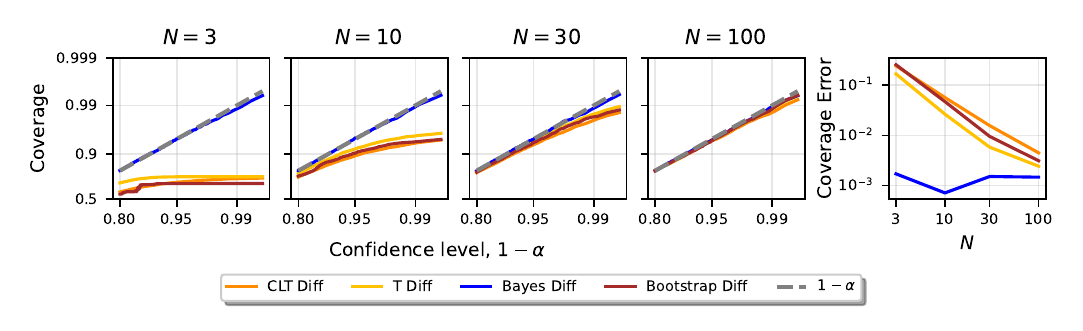}
        \vspace{-20pt}
        \caption{Coverage vs confidence level.}
        \label{fig:t_dist_coverage}
    \end{subfigure}%
    % \vspace{-5pt}
    
    \begin{subfigure}[b]{0.90\linewidth}
        \centering
        \includegraphics[width=\linewidth]{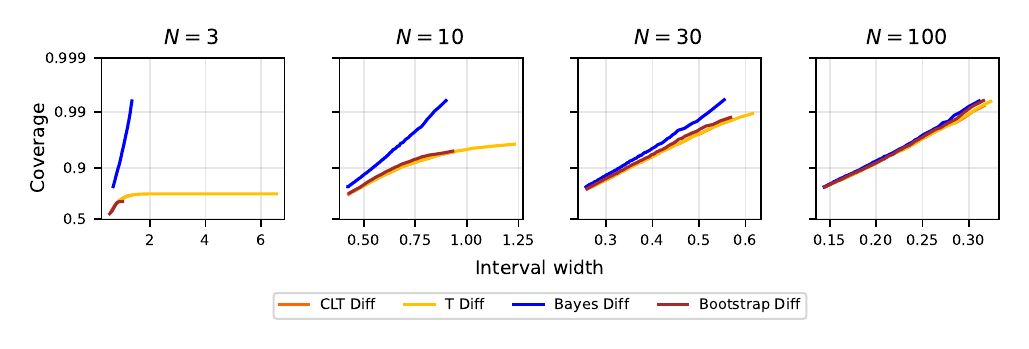}
        \vspace{-20pt}
        \caption{Coverage vs size of the intervals.}
        \label{fig:t_dist_size}
    \end{subfigure}
    % \vspace{-15pt}
    \caption{\textbf{Independent model comparison setting}. Intervals for the difference in means $\theta_A - \theta_B$.}
    \label{fig:t_dist_plot}
\end{figure}

Recall that in  \autoref{sec:clt}, we relied on the Slutsky's theorem to deal with the unknown variance. 
For smaller sample sizes, if the two assumptions above are approximately satisfied, the t-based intervals can have better properties than z-based ones. 
However, for the binary LLM evals we consider here, the variance $S^2$ is not independent of the sample mean. 
Nevertheless, in \autoref{fig:t_dist_plot} we show the properties of a t-based interval for the independent model comparison setting, where we construct intervals on the difference in means $\theta_A - \theta_B$.

\autoref{fig:t_dist_coverage} shows more favourable coverage properties compared to the CLT-based intervals. 
However, these intervals are extremely wide (\autoref{fig:t_dist_size}) with width exceeding 1 for small $N$. 
Given that the interval $[0, 1]$ achieves 100\% coverage for any $N$, the t-based intervals are clearly not useful. 
% potential references: \cite{reiczigel2003confidence, wallis2013binomial}

% Here we present 

\subsection{Wilson Score Intervals} \label{app:wilson_description}

The Wilson score  (WS) interval \citep{wilson1927probable} is applicable when constructing CIs for a single model's accuracy. It has improved coverage properties over the CLT approximation, particularly with small sample sizes. 

Given  Bernoulli data $y_i \iid \text{Bernoulli}(\theta)$, $i = 1,\ldots,N$, the WS interval is: 

\begin{equation}\label{eq:wilson}
    \CI_{1-\alpha, \text{Wilson}}(\theta) = \frac{\hat{\theta} + \frac{\zHalfAlpha^2}{2N}}{1 + \frac{\zHalfAlpha^2}{N}} \pm \frac{\frac{\zHalfAlpha}{2N}}{1 + \frac{\zHalfAlpha^2}{N}}\sqrt{4N\hat{\theta}(1 - \hat{\theta}) + \zHalfAlpha^2}
\end{equation}
where $\zHalfAlpha$ is the $\nicefrac{\alpha}{2}$-th quantile of the standard normal distribution.
(Note that the centre of the interval is no longer the sample mean $\hat{\theta}$---this helps to avoid the interval from collapsing to zero-width or extending past $[0,1]$.)

To arrive at this, we first take a normal approximation of the binomial with the sample standard deviation given by $\sqrt{\theta(1-\theta)/N}$:
\begin{equation}
    \zHalfAlpha \approx \frac{\theta - \hat{\theta}}{\sqrt{\theta(1-\theta)/N}}.
\end{equation}
Rearranging this we get a quadratic in $\theta$
\begin{align}
    \theta(1-\theta)\zHalfAlpha^2 &= (\theta - \hat{\theta})^2 \\
    0 &= (N + \zHalfAlpha^2) \theta^2 - (2N\hat{\theta} + \zHalfAlpha^2)\theta + N\hat{\theta}^2
\end{align}
which we can solve using the standard quadratic formula to find the upper and lower values of $\theta$ for the $1-\alpha$ confidence interval
\begin{align}
    p &= \frac{(2N\hat{\theta} + \zHalfAlpha^2) \pm \sqrt{(2N\hat{\theta} + \zHalfAlpha^2)^2 - 4(N + \zHalfAlpha^2)(N\hat{\theta}^2)}}{2(N + \zHalfAlpha^2)} \\
    &= \frac{2N\hat{\theta} + \zHalfAlpha^2}{2(N + \zHalfAlpha^2)} \pm \frac{\sqrt{4N^2
    \hat{\theta}^2 + 4N\hat{\theta}\zHalfAlpha^2 + \zHalfAlpha^4 - 4N^2\hat{\theta}^2 - 4N\hat{\theta}^2\zHalfAlpha^2}}{{2(N + \zHalfAlpha^2)}} \\
    &= \frac{2N\hat{\theta} + \zHalfAlpha^2}{2(N + \zHalfAlpha^2)} \pm \frac{\zHalfAlpha\sqrt{4N\hat{\theta} - 4N\hat{\theta}^2 + \zHalfAlpha^2}}{{2(N + \zHalfAlpha^2)}} \\
    &= \frac{2N\hat{\theta} + \zHalfAlpha^2}{2(N + \zHalfAlpha^2)} \pm \frac{\zHalfAlpha\sqrt{4N\hat{\theta}(1 - \hat{\theta}) + \zHalfAlpha^2}}{2(N + \zHalfAlpha^2)} \\
    &= \frac{\hat{\theta} + \frac{\zHalfAlpha^2}{2N}}{1 + \frac{\zHalfAlpha^2}{N}} \pm \frac{\frac{\zHalfAlpha}{2N}}{1 + \frac{\zHalfAlpha^2}{N}}\sqrt{4N\hat{\theta}(1 - \hat{\theta}) + \zHalfAlpha^2}.
\end{align}
Thus arriving at \autoref{eq:wilson}.

\subsection{Clopper-Pearson Intervals} \label{app:clopper_pearson_description}

Similarly to WS, the Clopper-Pearson (CP) interval \citep{clopper1934use} is  applicable when constructing CIs for a single model's accuracy. 
It is an \emph{exact} method based on the cumulative binomial distribution, albeit often yielding conservative (wider) intervals.

Given  Bernoulli data $y_i \iid \text{Bernoulli}(\theta)$, $i = 1,\ldots,N$, % is often referred to as an `exact' method.
%This is because it 
the CP interval is defined as containing all values of $\theta$ for which a two-sided binomial hypothesis test with significance level $\alpha$ does \textit{not} reject the null hypothesis $H_0: \theta = \hat{\theta}$ in favour of the alternative $H_1 : \theta \neq \hat{\theta}$, resulting in a coverage that is guaranteed to be at least $1-\alpha$.
We may write the CP interval as 
\begin{equation}\label{eq:clopper_pearson}
    \CI_{1-\alpha, \text{CP}}(\theta) = [\theta_\text{lower}, \theta_\text{upper}],
\end{equation}
where $\theta_\text{lower}$ and $\theta_\text{upper}$ are such that (denoting $\bar{y} = \frac{1}{N}\sum_{i=1}^N y_i$)
\begin{align}
    \sum_{k=N\bar{y}}^N \binom{N}{k} \theta_\text{lower}^k (1-\theta_\text{lower})^{n-k} &= \frac{\alpha}{2} \\
    \sum_{k=0}^{N\bar{y}} \binom{N}{k} \theta_\text{upper}^k (1-\theta_\text{upper})^{n-k} &= \frac{\alpha}{2}.
\end{align}
It can be shown \citep{thulin2014cost} that the values of $\theta_\text{lower}$ and $\theta_\text{upper}$ are given by
\begin{align}
    \theta_\text{lower} &= B\left(\frac{\alpha}{2}, \sum_{i=1}^N y_i, 1+\sum_{i=1}^N(1-y_i)\right) \\
    \theta_\text{upper} &= B\left(1-\frac{\alpha}{2}, 1+ \sum_{i=1}^N y_i, \sum_{i=1}^N(1-y_i)\right)
\end{align}
where $B(\alpha, a, b)$ is the $\alpha$-th quantile of the $\betadist(a,b)$ distribution.

\subsection{Delta Method}
As discussed in \autoref{sec:metrics_not_averages}, in some settings the metric we are interested in is not $\theta$ itself, but some value $g(\theta)$ (such as the $F_1$-score).
The usual way to estimate $g(\theta)$ is to use the \emph{plug-in estimator} $g(\hat\theta)$, where $\hat\theta$ is the maximum likelihood estimate of $\theta$.
In this case, the CLT cannot be directly applied to obtain confidence intervals for $g(\theta)$. 

Under standard regularity conditions, the delta method~\citep{oehlert1992note} can be used in conjunction with the CLT to approximate the sampling distribution of such metrics. 
By the (multivariate) CLT we have 
\begin{equation}
    \sqrt{n}(\hat{\theta} - \theta^\star) \xrightarrow{d} \mathcal{N}(0, \Sigma)
\end{equation}

Let $g: \theta \mapsto g(\theta)$ be a differentiable function. 
Then 
\begin{equation}
    \sqrt{n}(g(\hat{\theta}) - g(\theta)) \xrightarrow{d} \mathcal{N}(0, \nabla g(\theta)^\top \Sigma \nabla g(\theta)). \label{eq:delta_method}
\end{equation}

This allows us to construct an approximate $(1-\alpha)$ confidence interval for $g(\theta)$ as follows:
\begin{equation}
    \CI_{1-\alpha}(g(\theta)) =  g(\hat{\theta}) \pm z_{\alpha/2} \sqrt{\frac{1}{n} \nabla g(\hat{\theta})^\top \Sigma \nabla g(\hat{\theta})}.
\end{equation}
As before, we can rely on Slutsky's theorem and use the sample covariance $\hat \Sigma$ if the population covariance is unknown.

\subsection{Delta method for the $F_1$ score}

Let $\bm \theta = (\theta_\mathrm{TP}, \theta_\mathrm{FP}, \theta_\mathrm{FN}, \theta_\mathrm{TN})^\top$. The $F_1$ score can be written as
\begin{equation}
    g(\bm \theta) = F_1 =  \frac{2 \theta_\mathrm{TP}}{2\theta_\mathrm{TP} + \theta_\mathrm{FP} + \theta_\mathrm{FN}} = \frac{2 \theta_\text{TP}}{d} \quad d \coloneqq 2\theta_\mathrm{TP} + \theta_\mathrm{FP} + \theta_\mathrm{FN}.
\end{equation}
% The empirical 

% \begin{equation}
%     \hat F_1 = \frac{2 N_\mathrm{TP}}{2N_\mathrm{TP} + N_\mathrm{FP} + N_\mathrm{FN}} =
% \end{equation}

Then we have for the gradient $\nabla g(\bm \theta)$:
\begin{align}
    \nabla g(\bm\theta) &= 
    \begin{pmatrix}
\displaystyle  \frac{\partial g}{\partial \theta_{\mathrm{TP}}}\\[1.1em]
\displaystyle  \frac{\partial g}{\partial \theta_{\mathrm{FP}}}\\[1.1em]
\displaystyle  \frac{\partial g}{\partial \theta_{\mathrm{FN}}}\\[1.1em]
\displaystyle  \frac{\partial g}{\partial \theta_{\mathrm{TN}}}
\end{pmatrix} 
=\begin{pmatrix}
\displaystyle \frac{2(\theta_{\mathrm{FP}}+\theta_{\mathrm{FN}})}{d^2}\\[1.1em]
\displaystyle  -\frac{2\theta_{\mathrm{TP}}}{d^2}\\[1.1em]
\displaystyle  -\frac{2\theta_{\mathrm{TP}}}{d^2}\\[1.1em]
\displaystyle  0
\end{pmatrix}
=\begin{pmatrix}
\displaystyle \frac{2(1 - F_1)}{d}\\[1.1em]
\displaystyle -\frac{F_1}{d}\\[1.1em]
\displaystyle -\frac{F_1}{d}\\[1.1em]
\displaystyle0
\end{pmatrix}
\end{align}

For the covariance of $\bm \theta$ we have:
\begin{align}
    \Cov (\bm \theta) = \frac{1}{N}\,\Big(\text{diag}(\bm \theta) - \bm\theta\bm\theta^\top\Big)
\end{align}

The sample estimates of $\bm \theta$, $\Cov(\bm\theta)$ and $F_1$ are:
\begin{align}
    \hat {\bm \theta} &= \left(\dfrac{N_{\mathrm{TP}}}{N}, 
\frac{N_{\mathrm{FP}}}{N}, 
\frac{N_{\mathrm{FN}}}{N},
\frac{N_{\mathrm{TN}}}{N}\right)^\top \\
    \Cov (\bm \theta) &=  \frac{1}{N}
\Bigl(\text{diag}\bigl(\hat{\bm \theta}\bigr)
-\hat{\bm \theta}\,\hat{\bm \theta}^\top\Big) \\
    \hat{F}_1 & = \frac{2 N_\mathrm{TP}}{2N_\mathrm{TP} + N_\mathrm{FP} + N_\mathrm{FN}}
\end{align}
which we use to get the sample variance of $g(\bm \theta) = F_1$ as per \autoref{eq:delta_method}:
\begin{align}
    \Var(F_1) &= \nabla g(\bm \theta)^\top \, \Sigma \, \nabla g(\bm \theta) \\
    & \approx \nabla g(\hat{\bm\theta})^\top \Cov(\hat{\bm\theta})\, \nabla g(\hat{\bm\theta}) \\
    &=  \frac{1}{N}\, \nabla g(\hat{\bm\theta})^\top
\bigl[\text{diag}(\hat{\bm\theta})-\hat{\bm\theta}\,\hat{\bm\theta}^\top\bigr]
\nabla g(\hat{\bm\theta}) \\
&=
\frac{\hat{F}_1\,(1 - \hat{F}_1)\,(2 - \hat{F}_1)}{N\,d} \\
&=
\frac{\hat{F}_1\,(1 - \hat{F}_1)\,(2 - \hat{F}_1)}
     {2\,N_{\rm TP} + N_{\rm FP} + N_{\rm FN}}
\end{align}

Thus the standard error is
\begin{equation}
\SE(\hat{F}_1) =\sqrt{\frac{\hat{F}_1(1 - \hat{F}_1)(2 - F_1)}{2\,N_{\rm TP}+N_{\rm FP}+N_{\rm FN}}}\,,
\end{equation}
and the approximate two-sided $100(1-\alpha)\%$ confidence interval is
\begin{equation}
\label{eq:f1-ci}
    \CI_{1-\alpha}(F_1) = \hat{F}_1 \;\pm\; z_{1-\alpha/2}\;\sqrt{\frac{F_1(1 - F_1)(2 - F_1)}{2\,N_{\rm TP} + N_{\rm FP} + N_{\rm FN}}}\,.
\end{equation}

\clearpage
\section{Real-World Eval Error Bars}\label{app:real_data}

\subsection{Full LangChain Eval Error Bars}\label{app:langchain_full}
\begin{figure*}[th]
    % \centering
    \includegraphics[width=0.95\linewidth]{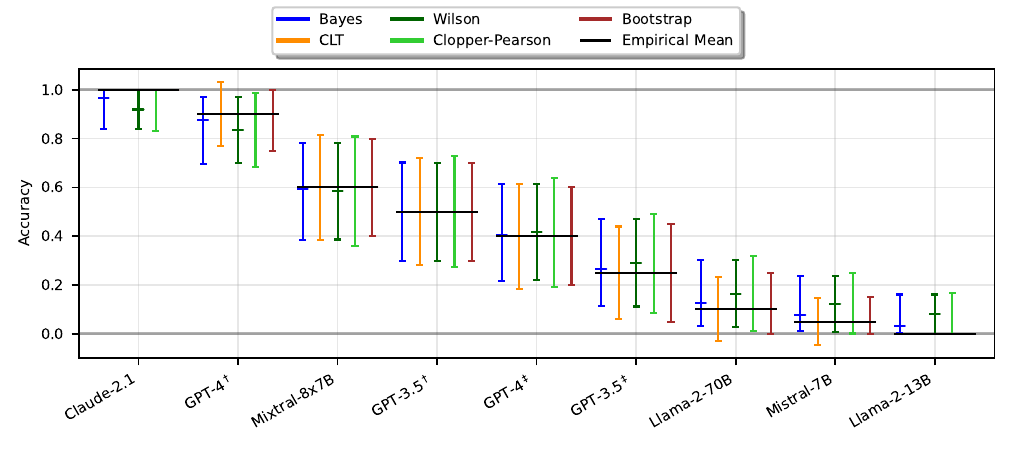}
    \caption{
        \textbf{Error bars on Langchain Tool-use Benchmark}. Extended version of \autoref{fig:real_data_langchain}. The benchmark consists of $N=20$ questions and we show 95\% confidence/credible intervals for the model accuracy, with the empirical mean shown in black.
    }
    \label{fig:real_data_langchain_full}
\end{figure*}

Here we present the error bars on all LLMs present in the Langchain dataset in \autoref{fig:real_data_langchain_full} for which we could find response data publicly on all $N=20$ questions\footnote{The raw evals data can be found along with code to reproduce all experiments in this paper at \url{https://github.com/sambowyer/no_clt_paper}.}.
% \footnote{The data can be found at \url{https://smith.langchain.com/public/128af05e-aa00-4e3b-a958-d166dd450581/d/compare?selectedSessions=17f86dec-f20b-445d-b97c-124a79a9e79a%2C0119bbdf-968b-4ceb-812a-3c0d2a619b8b%2C68866a1a-325a-41ac-a771-9150c1c2455f%2C1c885b09-212d-4608-bd3b-6e1b27964a88%2C59ef70f6-a0d2-4d3f-a1d9-265b9d736254%2Cc800fb51-7e88-43a7-a891-335271d48ced%2Cc12566e8-c6a4-4c8d-88f4-b026f03c787b%2C5b5b470f-8646-400f-b08e-a3559d2bb7d8%2C6e53b173-52a3-4d33-b691-3852fdc689a2%2Ca074b4f2-1758-4c0f-8bcb-fa142956835e%2C3aee9f81-2cb9-46f5-801b-c8f499eeba84&baseline=3aee9f81-2cb9-46f5-801b-c8f499eeba84}. \sam{not sure this is the correct link}}
The evals represent model success on Langchain's 26-tool typewriter task, in which LLM agents must spell out strings of letters by using 26 tools which each represent a letter of the alphabet.
Below, we clarify the more specific names of some of the LLMs in the figure, with the GPT models marked by a $^\dagger$ being the ones shown in \autoref{fig:real_data_langchain}.
\begin{itemize}
    % \item Claude 2.1: \verb|claude-2.1|
    \item GPT-4\(^\dagger\): \verb|gpt-4-1106-preview|
    \item Mixtral-8-7B: \verb|mixtral-8x7b-instruct|
    \item GPT-3.5\(^\dagger\): \verb|gpt-3.5-turbo-0613-openai|
    \item GPT-4\(^\ddagger\):  \verb|gpt-4-0613|
    \item GPT-3.5\(^\ddagger\):  \verb|gpt-3.5-turbo-1106|
    \item Llama-2-70B: \verb|llama-v2-70b-chat|
    \item Mistral-7B: \verb|mistral-7b-instruct|
    \item Llama-2-13B: \verb|llama-v2-13b-chat|
\end{itemize}

\clearpage
\subsection{Math Arena AIME 2025 II Error Bars}
\begin{figure*}[th]
    % \centering
    \includegraphics[width=0.95\linewidth]{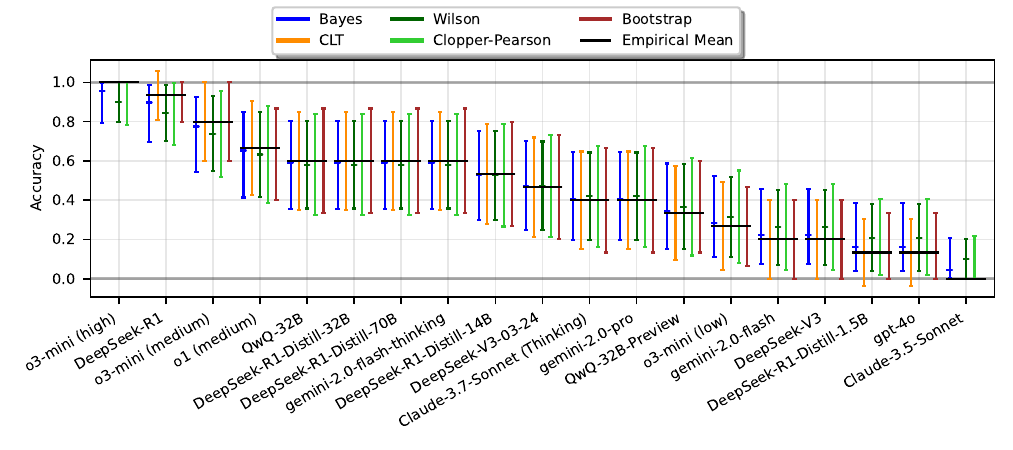}
    \caption{
        Error bars at 95\% confidence level computed for all models available on the Math Arena AIME 2025 II Benchmark ($N=15$).
    }
    \label{fig:real_data_matharena_full}
\end{figure*}

In \autoref{fig:real_data_matharena_full} we present error bars for all models on the Math Arena AIME 2025 II benchmark \citep{balunovic_srimatharena_2025, aops2025aime}.
We take each model's first attempt at each of the $N=15$ questions and compute 95\% confidence/credible intervals using the methods laid out in \autoref{sec:failure_simple_ci}.
Note that we again observe CLT-based error bars collapsing to zero-width (as do bootstrap error bars) and extending past $[0,1]$.

\clearpage
\section{Simulating Correlated Paired-Model Eval Data}\label{app:paired_simulation}

% \begin{wrapfigure}[16]{r}{0.33\textwidth}
% \vspace{-12pt}
%     \centering
%     \includegraphics[width=1.0\linewidth]{fig/rho_pdf.pdf}
%             \vspace{-18pt}
% \caption{\textbf{Distribution of correlation coefficient $\rho$.}}
%         \label{fig:rho_pdf} 
% \end{wrapfigure}
% \vspace{-20pt}

\begin{figure}
    \centering
    \begin{subfigure}[t]{0.45\textwidth}
        \centering
        \includegraphics[width=0.75\linewidth]{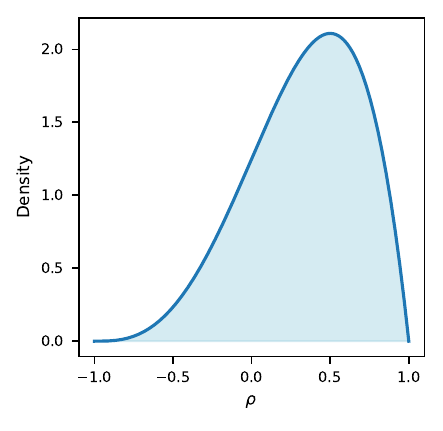}
        \caption{\textbf{Distribution of correlation coefficient $\rho$.}}
        \label{fig:rho_pdf} 
    \end{subfigure}
    \hfill
    \begin{subfigure}[t]{0.45\textwidth}
        \centering
        \includegraphics[width=0.75\linewidth]{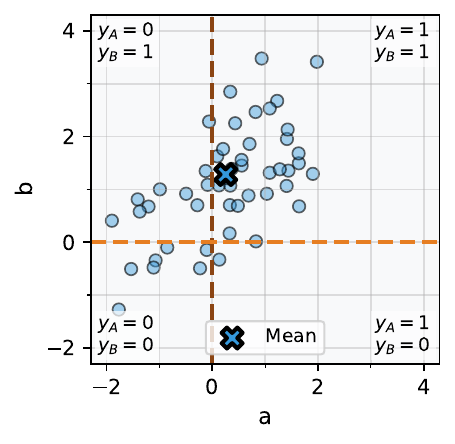}
        \caption{\textbf{Paired model data generation illustration} with  $\rho = 0.7$, $\theta_A=0.6$, $\theta_B = 0.9$ and $N=50$ samples.}
        \label{fig:paired_dgp}
    \end{subfigure}
    \caption{}%\textbf{(a)} Distribution of correlation coefficient $\rho$. \textbf{(b)} Paired model data generation illustration with  $\rho = 0.7$, $\theta_A=0.6$, $\theta_B = 0.9$ and $N=50$ samples.}
\end{figure}

In \autoref{sec:paired}, we wanted to generate pairs of eval results for two LLMs, $A$ and $B$, such that the responses between the two models were correlated.
In order to maintain control on the marginal values of $\theta_A$ and $\theta_B$, we sample these uniformly in the same way as in the rest of the paper:
\begin{align}
\theta_A, \theta_B \iid \betadist(1,1) = \uniform[0,1].
\end{align}
However, to induce correlation between evals we clearly can't just use these values as Bernoulli parameters independently.
First, we sample  a correlation coefficient $\rho = 2\hat{\rho} - 1$ where $\hat{\rho} \sim \betadist(4,2)$, leading to the distribution of $\rho$ shown in \autoref{fig:rho_pdf}.
The preference for positive correlations follows on from real-world intuition: we'd expect results from two LLMs to be positively correlated much more of the time than negatively correlated.

Next, we sample $N$ points, $\{(a_i,b_i)\}_{i=1}^N$, on a bivariate Gaussian with unit variances and correlation coefficient $\rho$:
\begin{align}
(a_i,b_i) &\iid \mathcal{N}\left((\mu_A \quad \mu_B)^\top, \Sigma\right), \\
\mu_A &= \Phi^{-1}(\theta_A), \\
\mu_B &= \Phi^{-1}(\theta_B), \\
\Sigma &= \begin{pmatrix}1 & \rho \\ \rho & 1\end{pmatrix},
\end{align}
where $\Phi$ is the standard univariate Gaussian CDF.

Since the marginal distributions of $a_i$ and $b_i$ are standard Gaussians centred at $\mu_A$ and $\mu_B$ respectively, we have that $(a_i - \mu_A), (b_i - \mu_B) \sim \mathcal{N}(0,1)$ and therefore
\begin{align}
\mathbb{P}(a_i > 0) &= 1 - \mathbb{P}(a_i - \mu_A < -\mu_A) = 1 - \Phi(-\mu_A) = \Phi(\mu_A) = \theta_A, \\
\mathbb{P}(b_i > 0) &= 1 - \mathbb{P}(b_i - \mu_B < -\mu_B) = 1 - \Phi(-\mu_B) = \Phi(\mu_B) = \theta_B.
\end{align}
% \begin{wrapfigure}[16]{r}{0.33\textwidth}
% \vspace{-12pt}
%     \centering
%     \includegraphics[width=1.0\linewidth]{fig/joint_distribution_latents.pdf}
%     \vspace{-18pt}
%     \caption{\textbf{Paired model data generation illustration} with  $\rho = 0.7$, $\theta_A=0.6$, $\theta_B = 0.9$ and $N=50$ samples.}
%     \label{fig:paired_dgp}
% \end{wrapfigure}
Hence, if we assign binary eval outcomes for both LLMs according to the signs of $a_i$ and $b_i$ for $i = 1, \ldots, N$, we arrive at Bernoulli marginals, as desired:
\begin{align}
    y_{A;i} &= \mathbbm{1}[a_i > 0],\\
    y_{B;i} &= \mathbbm{1}[b_i > 0].
\end{align}
The choice of the covariance matrix $\Sigma$ ensures that the evals for the two LLMs are are correlated with the desired level of correlation $\rho$.

We illustrate this data generation procedure in \autoref{fig:paired_dgp}. 
We generate a paired eval dataset with a positive correlation of $\rho = 0.7$ between the two models, A and B. 
The success probabilities are $\theta_A = 0.6$ and $\theta_B = 0.9$. 
The points in the scatter plot are samples from a bivariate Gaussian distribution with mean $\big(\Phi^{-1}(\theta_A)\quad \Phi^{-1}(\theta_B)\big)^\top=(0.25 \quad 1.28)^\top$ (denoted with a blue cross), unit variance and correlation $0.7$. 
The threshold lines at $a = 0$ (dashed orange horizontal line) and $b = 0$ (dashed brown vertical line) divide the space into four quadrants, each corresponding to a different combination of binary outcomes $y_A$ and $y_B$, as labelled in the corners. 
% The empirical success probabilities for this binary eval are $\hat{\theta}_A=66\%$ and $\hat{\theta}_B=84\%$

See Appendix \ref{app:paired_bayes_code} for details on how inference is done in this model via importance sampling.

\section{Pass@K Metric}
In \autoref{sec:metrics_not_averages} we examined $F_1$ scores as a metric more immediately amenable to Bayesian than frequentist analysis, but many other metrics would also fit in here too. 
For example, consider the common pass@K metric, which reports the proportion of runs in which a model gives the correct answer within its first $K$ (IID) generations.

In the Bayesian setting, we can infer a Bernoulli posterior over the probability of a single generation on a single question being correct for a certain model.
This can then easily be used to compute error bars for the probability of \textit{at least} 1 in $K$ (IID) generations being correct.

However, since we're really interested in a single number that measures the performance of a model on all the questions in a dataset, we'd need a hierarchical Bayesian model with a per-model latent variable, which captures the average probability of a single generation being correct on a single, randomly chosen question.
This might look similar to the hierarchical clustered model from \autoref{sec:clt-clustered}, but with a hierarchy over models rather than tasks. 
This model could also be extended to have an hierarchy over both models and questions/tasks.
It is unclear to us how best to construct a corresponding CLT-based/frequentist interval.

\clearpage

\section{Python Code for Importance Sampling}
Here we provide simple code for performing the importance sampling mentioned in \autoref{sec:clt-clustered} and \autoref{sec:paired}.
For a review of importance sampling techniques, see \cite{tokdar2010importance}.

In general, importance sampling for Bayesian inference works by drawing some $K$ samples, $\{\theta^{(k)}\}_{k=1}^K$, of a latent variable $\theta$ from some proposal distribution $Q$ and computing their importance weight as the ratio between the likelihood under the generative model $P(Y,\theta)$ and the proposal $Q(\theta)$,
\begin{align}
    w^{(k)} = \frac{P(Y, \theta)}{Q(\theta)}
\end{align}
where $Y$ is our data.
Using Bayes' rule, we can rewrite this to show that if our proposal is simply the prior, $Q(\theta) = P(\theta)$, then the importance weights are directly given by the likelihood
\begin{align}
    w^{(k)} &= \frac{P(Y, \theta^{(k)})}{Q(\theta^{(k)})} \\
    &= \frac{P(Y | \theta^{(k)})P(\theta^{(k)})}{Q(\theta^{(k)})} \\
    &= P(Y | \theta^{(k)}).
\end{align}
Then we can, for example, approximate the posterior expectation of the latent $\theta$ using normalised versions of the importance weights
\begin{align}
    \hat{w}^{(k)} = \frac{w^{(k)}}{\sum_{i=1}^K w^{(i)}}
\end{align}
since
\begin{align}
    \mathbb{E}_{\theta \sim P(\cdot | Y)} [\theta] &= \int \theta P(\theta | Y)d\theta \\
    &= \int \theta \frac{P(Y | \theta)P(\theta)}{P(Y)}   d\theta \\
    &= \int \theta \frac{P(Y | \theta)Q(\theta)}{P(Y)}   d\theta \\
    &= \mathbb{E}_{\theta \sim Q(\cdot)} \left[ \theta\frac{P(Y | \theta)}{P(Y)} \right]\\
    &\approx \frac{1}{K}\sum_{k=1}^K \theta^{(k)} \hat{w}^{(k)}.
\end{align}
% The final approximation works because $\hat{w}^{(k)} \approx \frac{P(Y| \theta^{(k)})}{P(Y)}$.

\subsection{Importance Sampling in the Clustered Setting}\label{app:clustered_bayes_code}
The generative model for the clustered setting is as follows:
\begin{align}
  d &\sim \gammadist(1, 1) \\
  \theta &\sim \betadist(1, 1) \\
  \theta_t &\sim \betadist(d \theta, d (1-\theta)) \\
  y_{i,t} &\sim \bernoulli(\theta_t).
\end{align}
% were, $d$ controls the the range of difficulties of the tasks or cluster (i.e. its \emph{dispersion}), $\theta$ is the global performance of the model, and is the quantity we are trying to infer, while $\theta_t$ is the performance on a given task.  
To perform Bayesian inference on $\theta$ we integrate out $\theta_t$: 
\vspace{-8pt}
\begin{align}
  Y_t &\sim \betabin(N_t, d \theta, d (1-\theta)).
\end{align}
where $Y_t = \sum_{i=1}^{N_t} y_{i,t}$ is the total number of correct answers in task $t$.
As described in the main text, we use the prior as a proposal to obtain $K=10,000$ samples $\{(\theta^{(k)}, d^{(k)})\}_{k=1}^K$ which have an associated importance weight:
% Now there are only two latent variables, $\theta$ and $d$, meaning that the inference problem is fairly straightforward, and amenable to many different approaches.
% We use importance sampling with the prior as proposal: \(\theta \sim \betadist(1, 1)\) and \(d \sim \text{Gamma}(1,1)\).
% Then the importance weights $\{w^{(k)}\}_{k=1}^K$ are  given by the Beta-Binomial likelihood of the data $Y_t$ under each $(\theta^{(k)}, d^{(k)})$: 
\begin{align}
    w^{(k)} = \prod_{t=1}^T \betabin(Y_t; N_t, d^{(k)} \theta^{(k)}, d^{(k)}(1-\theta^{(k)})).
\end{align}
We then resample the set $\{\theta^{(k)}\}_{k=1}^K$ with repeats using the weights $\{w^{(k)}\}_{k=1}^K$ and report credible intervals by taking the relevant percentiles of the resulting set of posterior samples.
% (Note that the code below computes the log-weights and renormalises them in a numerically stable way.)

% This works since 
% \begin{equation}
%     \mathbb{P}(\theta|\{Y_t\}_{t=1}^T) = \mathbb{}
% \end{equation}

\begin{pbox}[label={ex:bayes_clustered}]{Bayesian analysis for clustered evals}
\begin{minted}{python}
# S_t, N_t: np.arrays of length T with total
# sucesses & questions per task
import numpy as np
from scipy.stats import betabinom

# set number of samples, K
K = 10_000

# get K samples from the prior (with extra dimension for broadcasting over tasks)
thetas = np.random.beta(1,1, size=(K,1))
ds = np.random.gamma(1,1, size=(K,1))

# obtain weights via the likelihood (sum the per-task log-probs)
log_weights = betabinom(N_t, (ds*thetas), (ds*(1-thetas))).logpmf(S_t).sum(-1)

# normalise the weights
weights = np.exp(log_weights - log_weights.max())
weights /= weights.sum()

# obtain samples from the posterior
posterior = thetas[np.random.choice(K, size=K, replace=True, p=weights)]

# Bayesian credible interval
bayes_ci = np.percentile(posterior, [2.5, 97.5])
\end{minted}
\end{pbox}

\subsection{Importance Sampling in the Paired Setting}\label{app:paired_bayes_code}
In the paired setting we use the following generative model, as shown in Appendix \ref{app:paired_simulation}:
\begin{align}
  \theta_A &\sim \betadist(1, 1) \\
  \theta_B &\sim \betadist(1, 1) \\
  \hat{\rho} &\sim \betadist(4,2) \\
  \rho &= 2\hat{\rho} - 1 \\
(a_i,b_i) &\iid \mathcal{N}\left(\mu, \Sigma\right), \\
    y_{A;i} &= \mathbbm{1}[a_i > 0],\\
    y_{B;i} &= \mathbbm{1}[b_i > 0],
\end{align}
for $i=1,\ldots,N$, such that
\begin{align}
\mu &= \begin{pmatrix}\Phi^{-1}(\theta_A) \quad \Phi^{-1}(\theta_B) \end{pmatrix}^T \\
\Sigma &= \begin{pmatrix}1 & \rho \\ \rho & 1\end{pmatrix},
\end{align}
where $\Phi$ is the standard univariate Gaussian CDF.

To perform Bayesian inference on $\theta_A - \theta_B$, we again use importance sampling where we draw $K=10,000$ samples from the prior as a proposal to obtain $\{(\theta_A^{(k)}, \theta_B^{(k)}, \rho^{(k)})\}_{k=1}^K$.
Then we calculate the importance weights as
\begin{align}
    \label{eq:paired_likelihood}
    w^{(k)} = \prod_{i=1}^N p(y_{A;i}, y_{B;i} | \theta_A^{(k)}, \theta_B^{(k)}, \rho^{(k)}). 
\end{align}
In order to calculate this quantity, we break the problem into four cases, based on the possible combinations of success/failure for model A and model B.
For a single question $i \in [N]$, we have
\begin{align}
    p(y_{A;i}, y_{B;i} | \theta_A^{(k)}, \theta_B^{(k)}, \rho^{(k)}) = \begin{cases}
        \theta_{AB}^{(k)} \defeq \mathbb{P}(a_i, b_i > 0) & \text{ if } a_i = b_i = 1, \\
        \theta_{AB^\bot}^{(k)} \defeq\mathbb{P}(a_i, > 0, b_i < 0) & \text{ if } a_i = 1 \text { and } b_i = 0 \\
        \theta_{A^\bot B}^{(k)} \defeq \mathbb{P}(a_i, < 0, b_i > 0) & \text{ if } a_i = 0 \text { and } b_i = 1 \\
        \theta_{A^\bot B^\bot}^{(k)} \defeq \mathbb{P}(a_i, b_i < 0) & \text{ if } a_i = b_i = 0. \\
    \end{cases}
\end{align}

If we can figure out the values of $\theta_{AB}^{(k)}, \theta_{A B^\bot}^{(k)}, \theta_{A^\bot B}^{(k)}$, and $\theta_{A^\bot B^\bot}^{(k)}$, then we just need the number of occurrences of each combination in the data in order to evaluate \autoref{eq:paired_likelihood}:
\begin{align}\label{eq:paired_likelihood_expanded}
    w^{(k)} &= \prod_{i=1}^N p(y_{A;i}, y_{B;i} | \theta_A^{(k)}, \theta_B^{(k)}, \rho^{(k)}) \\
    &= (\theta_{AB}^{(k)})^S (\theta_{A B^\bot}^{(k)})^T (\theta_{A^\bot B}^{(k)})^U (\theta_{A^\bot B^\bot}^{(k)})^V,
\end{align}
where
\begin{align}
    S &= \sum_{i=1}^N
    y_{A;i}y_{B;i}, \\
    T &= \sum_{i=1}^N
    y_{A;i}(1-y_{B;i}), \\
    U &= \sum_{i=1}^N
    (1-y_{A;i})y_{B;i}, \\
    V &= \sum_{i=1}^N
    (1-y_{A;i})(1-y_{B;i}).
\end{align}

Now, note that 
\begin{equation}
\theta_{A^\bot B^\bot}^{(k)} = \mathbb{P}(a_i^{(k)},b_i^{(k)} < 0) = \Phi_2(\bm{0},; \mu^{(k)}, \Sigma^{(k)}),
\end{equation}
where $\Phi_2$ is the bivariate Gaussian CDF (parameterised by $\mu^{(k)}$ and $\Sigma^{(k)})$), which can be calculated numerically.
Specifically, we adopt the simple approximation derived in \citet{Tsay03042023}, an implementation of which can be found at \url{https://github.com/sambowyer/bayes_evals}.

Knowing $\theta_{A^\bot B^\bot}^{(k)}$ allows you to calculate the other three probabilities, since we know the following:
\begin{align}
    1 &= \theta_{AB}^{(k)} + \theta_{A B^\bot}^{(k)} + \theta_{A^\bot B}^{(k)} + \theta_{A^\bot B^\bot}^{(k)}, \\
    \theta_A^{(k)} &= \theta_{AB}^{(k)} + \theta_{A B^\bot}^{(k)}, \\
    \theta_B^{(k)} &= \theta_{AB}^{(k)} + \theta_{A^\bot B}^{(k)}.
\end{align}

Thus, we can calculate the importance weights from \autoref{eq:paired_likelihood_expanded} with the following relationships:
\begin{align}
    \theta_{AB}^{(k)} &= \theta_{A^\bot B^\bot}^{(k)} + \theta_A^{(k)} + \theta_B^{(k)} - 1, \\
    \theta_{A B^\bot}^{(k)} &= 1 - \theta_B^{(k)} - \theta_{A^\bot B^\bot}^{(k)}, \\
    \theta_{A^\bot B}^{(k)} &= 1 - \theta_A^{(k)} - \theta_{A^\bot B^\bot}^{(k)}.
\end{align}

% \begin{align}
%     w^{(k)} = \prod_{i=1}^N \left(\theta_A^{(k)} y_{A;i} + (1-\theta_A^{(k)})(1-y_{A;i}) \right)\left(\theta_{B;i}^{(k)} y_{B;i} + (1-\theta_{B;i}^{(k)})(1-y_{B;i}) \right)
% \end{align}
% where $\theta_{B;i}$ is our per-question Bernoulli parameter for model B:
% \begin{align}
%     \theta_{B;i} = \theta_B + \lambda (y_{A; i}-\theta_A).
% \end{align}

As in the previous section, App. \ref{app:clustered_bayes_code}, we obtain a collection of $K$ posterior samples, $\{(\theta^{(k, \text{posterior})}_A, \theta^{(k, \text{posterior})}_B)\}_{k=1}^K$, by resampling the set $\{(\theta^{(k)}_A, \theta^{(k)}_B)\}_{k=1}^K$ with repeats using the weights $\{w^{(k)}\}_{k=1}^K$.
To calculate our credible intervals on the value of $\theta_A - \theta_B$, we take the relevant percentiles from the set $\{\theta^{(k, \text{posterior})}_A - \theta^{(k, \text{posterior})}_B\}_{k=1}^K$.

\begin{pbox}[label={ex:bayes_paired}]{Bayesian analysis for paired evals}
\begin{minted}{python}
# y_A, y_B: length N binary "eval" vectors
import numpy as np
from numpy.random import beta
from scipy.stats import norm
from binorm import binorm_cdf # 2D Gaussian CDF, defined elsewhere

# set number of samples, K
K = 10_000

# get K samples from the prior (with extra dimension for broadcasting over questions)
theta_As = beta(1,1, size=K)
theta_Bs = beta(1,1, size=K)
rhos     = 2*beta(4,2, size=K) - 1


# 2x2 contingency table (flattened)
S = (y_A * y_B).sum(-1)             # S = A correct,   B correct
T = (y_A * (1 - y_B)).sum(-1)       # T = A correct,   B incorrect
U = ((1 - y_A) * y_B).sum(-1)       # U = A incorrect, B correct
V = ((1 - y_A) * (1 - y_B)).sum(-1) # V = A incorrect, B incorrect

# calculate the bivariate normal mean
mu_As = norm(0,1).ppf(theta_As)
mu_Bs = norm(0,1).ppf(theta_Bs)

# Calculate probabilities of each cell in the 2x2 table
theta_V = binorm_cdf(x1=0, x2=0, mu1=mu_As, mu2=mu_Bs, sigma1=1, sigma2=1, rho=rhos)
theta_S = theta_As + theta_Bs + theta_V - 1
theta_T = 1 - theta_Bs - theta_V
theta_U = 1 - theta_As - theta_V

# Due to numerical issues, we need to handle the case where the probabilities are not in [0,1]
# (probabilities may be very small and negative instead of 0)
valid_idx = (theta_S > 0) & (theta_T > 0) & (theta_U > 0) & (theta_V > 0) 
log_weights = S*np.log(theta_S[valid_idx]) + T*np.log(theta_T[valid_idx]) + \
              U*np.log(theta_U[valid_idx]) + V*np.log(theta_V[valid_idx])

# normalise the weights
weights = np.zeros(K)
weights[valid_idx] = np.exp(log_weights - log_weights.max())
weights /= weights.sum()

# obtain samples from the posterior
posterior = (theta_As - theta_Bs)[np.random.choice(K, size=K, replace=True, p=weights)]

# Bayesian credible interval
bayes_ci = np.percentile(posterior, [2.5, 97.5])
\end{minted}
\end{pbox}

\clearpage

\section{Computational Cost}\label{app:compute_time}

\begin{figure}[htbp]
    \centering
    \includegraphics[width=\linewidth]{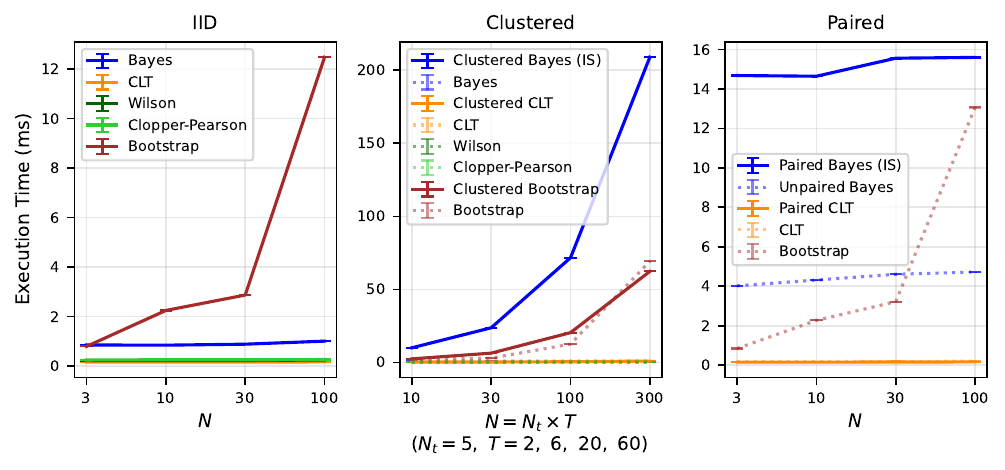}
    \caption{
    \textbf{Computational time.}
        Time (in milliseconds) required to compute error bars on a set of $N$ evals with different methods. 
        Results are averaged over 1000 repeats on a single CPU, and error bars (which are close to zero) report standard errors. 
    }
    \label{fig:combined_timing}
\end{figure}

In \autoref{fig:combined_timing} we report the mean computation time of the different methods discussed in \autoref{sec:clt_ci}, \autoref{sec:clt-clustered}, and \autoref{sec:paired} averaged over 1000 repeated runs on a single CPU.

Whilst we observe the Bayesian (and bootstrap) methods taking longer than the CLT-based methods, in the few-data setting of this paper the computational cost of all of these methods is trivial. 
(The longest compute-time was 200 milliseconds for the largest Bayesian model in \autoref{fig:clustered_intervals}.)
In the case where $N$ is much larger and compute cost starts to grow, the faster CLT-based methods would perform acceptably.

\clearpage

\section{Robustness of Experiments to Random Seeding} \label{app:error_bars_on_error_bars}

\begin{figure*}[th]
    \centering
    \includegraphics[width=0.90\linewidth]{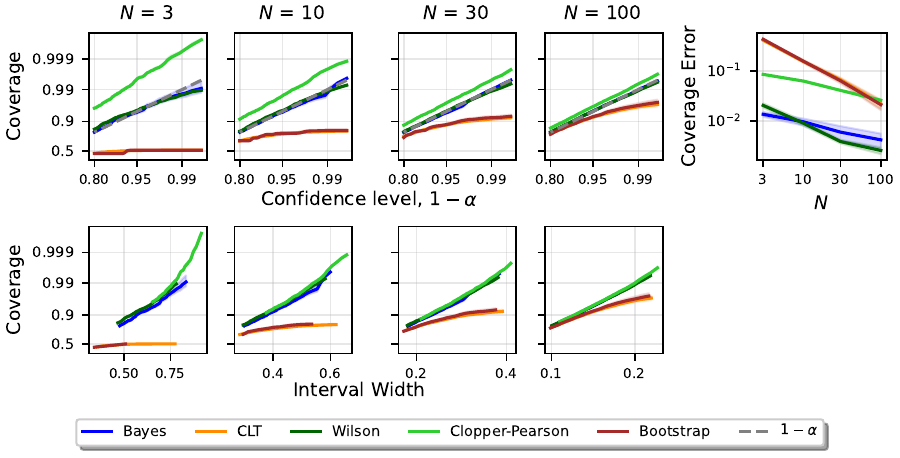}
    \vspace{-8pt}
    \caption{\textbf{IID question setting.} Coverage vs. confidence level (top) and coverage vs. width (bottom) for various methods. Coverage error denotes the mean absolute difference between true and nominal coverage for each method at a given value of $N$. \textbf{Error bars are given as faint lines around the means representing the standard error over 5 repeated experiments.} These error bars are very small, suggesting that each experiment successfully extracts values very close to true coverage per method.}
    \label{fig:error_bars_on_error_bars}
\end{figure*}

In \autoref{fig:error_bars_on_error_bars} we show an alternative version of \autoref{fig:iid_intervals} (comparing various methods in the IID setting of \autoref{sec:failure_simple_ci}) in which results are averaged over five runs using different random seeds.
Error bars representing standard errors are shown in faint colours but are generally very small (on the order of $10^{-3}$ or smaller) and therefore often hard to identify.
This suggests that the experiment methodology is robust to the randomness inherent in the data generating procedure.

\clearpage
\section{Interval Width vs. Coverage Plots}\label{app:interval_width}
\begin{figure*}[th]
    \centering
    \includegraphics[width=0.90\linewidth]{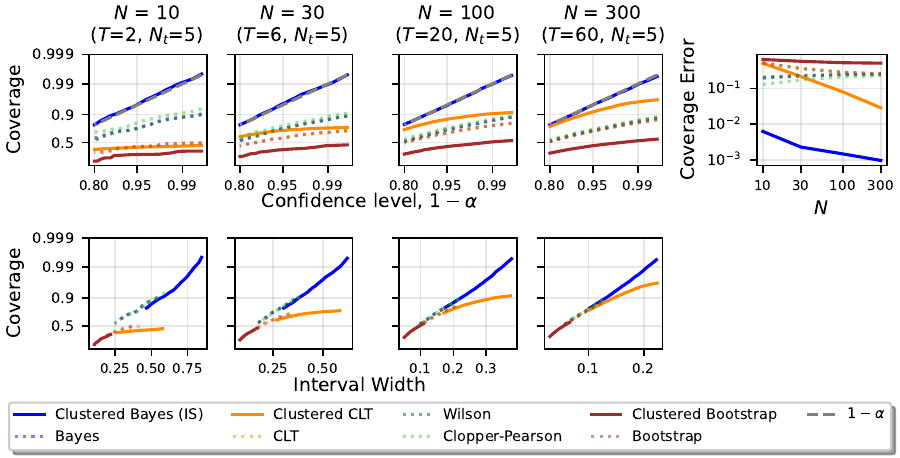}
    \vspace{-8pt}
    \caption{\textbf{Clustered questions setting}. Coverage vs. confidence level (\textbf{top}) and vs. interval-width (\textbf{bottom}) for various interval-calculation methods on the value of $\theta$.
    Methods ignoring the clustered structure of the data---assuming instead IID questions as per \autoref{sec:failure_simple_ci}---are shown as dotted lines.
    Results are averaged over 100 values of $\theta \sim \betadist(1,1)$, each with 200 repeated experiments with randomly generated datasets.}
    \label{fig:clustered_intervals_with_width}
\end{figure*}

\begin{figure}
    \centering
    \includegraphics[width=0.90\linewidth]{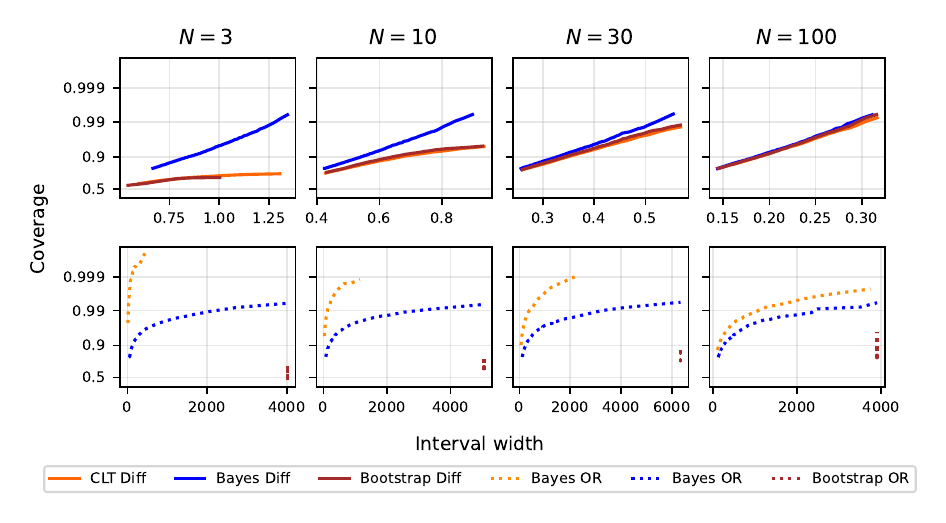}
    \vspace{-15pt}
    \caption{ \textbf{Independent model comparison setting}. Coverage vs. interval-width for various interval-calculation methods.
    The top row includes intervals for the difference, $\theta_A - \theta_B$, whilst the bottom row shows intervals for the odds ratio.
    %100 values of $\theta \sim \betadist(1,1)$, each with 200 repeated experiments with randomly generated datasets.
    Note that in the odds ratio methods, we have dropped infinite width values for the method based on the Fisher exact test, which accounted for 43.5\%,  17\%, 6.5\% and 1.9\% for $N=3, 10, 30 \text{ and } 100$, respectively. \
    We have also clipped bootstrap widths as it produced extremely large (order $1e14$, essentially infinite) intervals.
    % Infinite values are dropped for the fisher; clipped Bootstrap as it's producing huge numbers as well.}
    }
    \label{fig:indep_comparison_sizes}
\end{figure}

\begin{figure*}[t]
    \centering
    \includegraphics[width=0.90\linewidth]{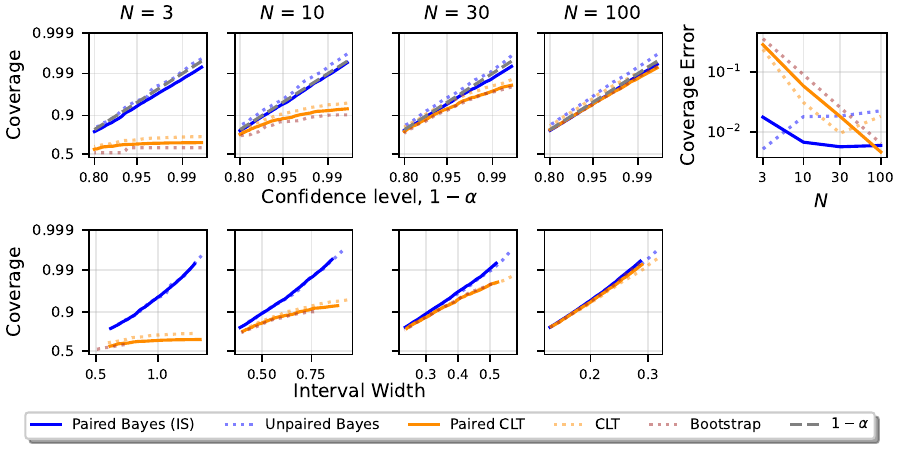}
    \vspace{-15pt}
    \caption{\textbf{Paired questions setting}. Coverage vs. confidence level (\textbf{top}) and vs. interval-width (\textbf{bottom}) for various interval-calculation methods on the value of $\theta_A - \theta_B$.
    Methods ignoring the paired structure of the data---assuming instead IID questions and answers from model A and from model B, as per \autoref{sec:failure_simple_ci}---are shown as dotted lines.
    Results are averaged over 100 values of $\theta_A, \theta_B \sim \betadist(1,1)$, each with 200 repeated experiments with randomly generated datasets.}
    \label{fig:paired_intervals_with_width}
\end{figure*}

Here we present expanded versions of \autoref{fig:clustered_intervals}, \autoref{fig:indep_model_comparison} and \autoref{fig:paired_intervals} which include the original coverage vs. confidence level results in their top rows as well as a bottom row showing coverage vs. interval width.
\autoref{fig:clustered_intervals_with_width} presents these results for the clustered question setting of \autoref{sec:clt-clustered}, \autoref{fig:indep_comparison_sizes} shows the independent model comparison setting of \autoref{sec:indep_model_comparison}, and \autoref{fig:paired_intervals_with_width} shows the paired question setting of \autoref{sec:paired}.

In each case, we observe that, much like in \autoref{sec:failure_simple_ci}, the Bayesian methods tend to generate much more efficient (i.e. narrower) error bars for a given coverage than CLT- and bootstrap-based methods.

\clearpage
\section{Ablations}\label{app:ablations}

% If our data truly comes from the assumed sampling distribution (likelihood) and our chosen prior distribution accurately reflects the parameter's true distribution, then credible intervals constructing using the true posterior distribution will correctly capture the true conditional probability distribution of the parameter given the observed data. 
% This means that the credible intervals we get will achieve the nominal coverage, $(1 - \alpha)$ that we choose, for example, a 95\% credible interval will contain the true parameter value 95\% of the time.

% [we ablate the prior specification in such and such way.]

% In order to evaluate fully the behaviour of each interval-generating method, we present results in which the true prior over model performance is not uniform, as is assumed by the Bayesian methods

\subsection{IID Questions Setting}\label{app:iid_mismatch}
\begin{figure}
    \centering
    \includegraphics[width=\linewidth]{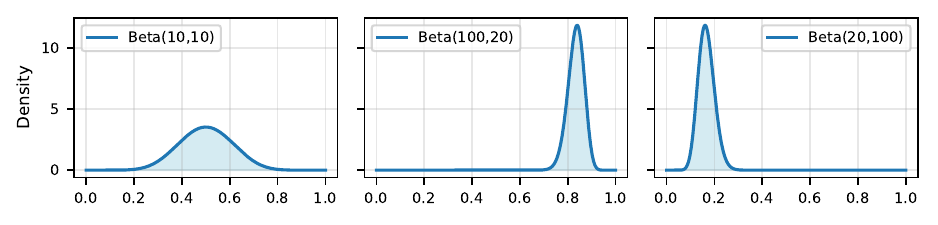}
    \caption{
        \textbf{Probability density functions}. \(\betadist(10,10)\) \textbf{(left)};  \(\betadist(100,20)\)  \textbf{(centre)}; \(\betadist(20,100)\) \textbf{(right)}.
    }
    \label{fig:beta-pdfs} 
\end{figure}

\begin{figure}[htbp]
    \centering
    \includegraphics[width=0.9\linewidth]{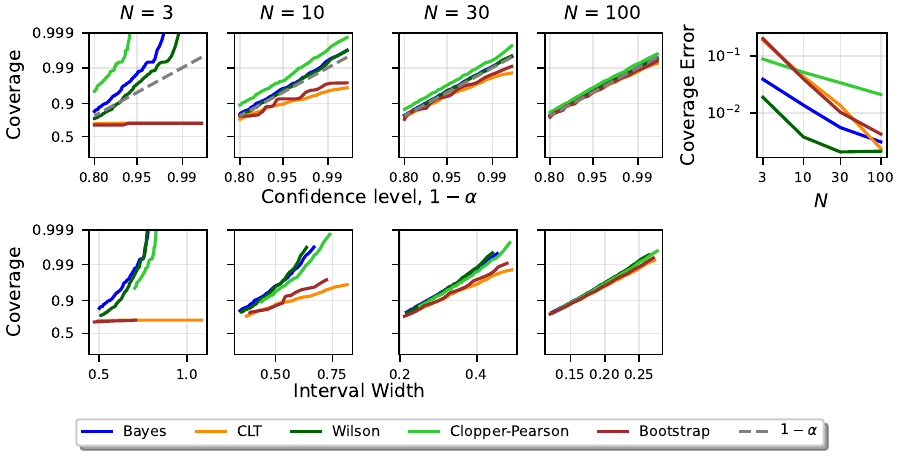}
    \vspace{-5pt}
    \caption{
        \textbf{IID question setting.} Coverage of intervals on \(\theta\) with mismatched \(\theta \sim \betadist(10,10)\) prior.
    }
    \label{fig:iid_intervals_beta_10_10}
\end{figure}

\begin{figure}[htbp]
    \centering
    \includegraphics[width=0.9\linewidth]{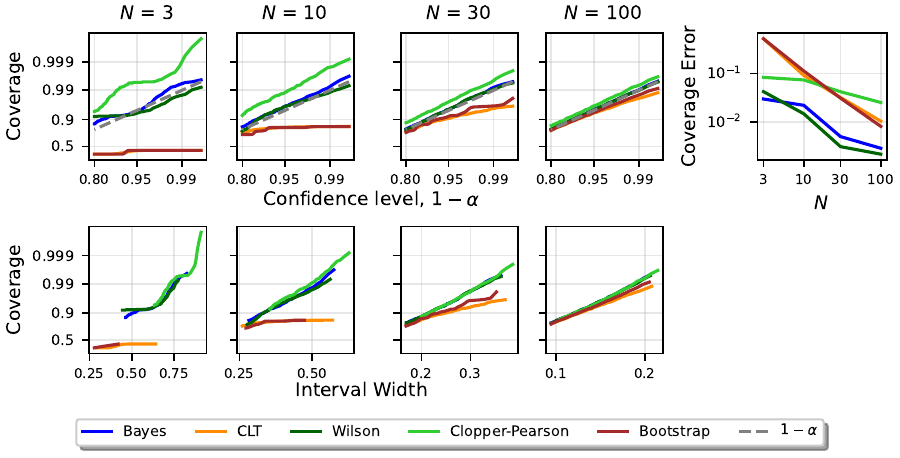}
    \vspace{-5pt}
    \caption{
       \textbf{IID question setting.} Coverage of intervals on \(\theta\) with mismatched \(\theta \sim \betadist(100,20)\) prior.
    }
    \label{fig:iid_intervals_beta_100_20}
\end{figure}

\begin{figure}[htbp]
    \centering
    \includegraphics[width=0.9\linewidth]{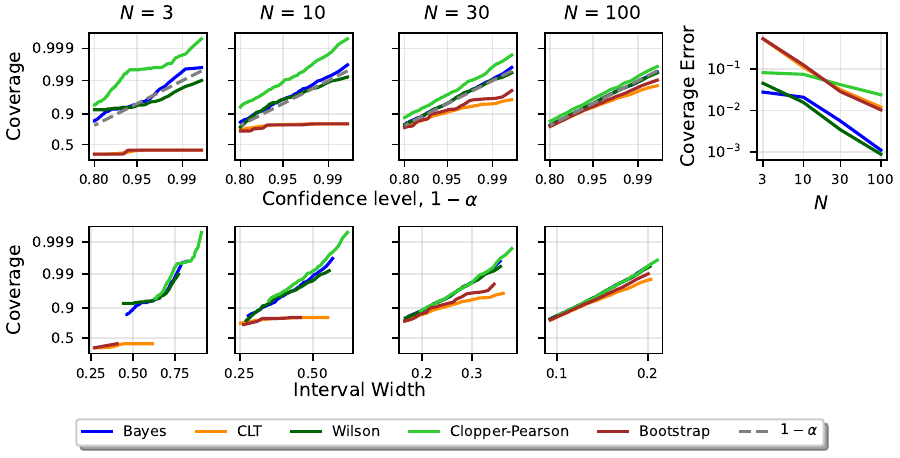}
    \vspace{-5pt}
    \caption{
       \textbf{IID question setting.} Coverage of intervals on \(\theta\) with mismatched \(\theta \sim \betadist(20,100)\) prior.
    }
    \label{fig:iid_intervals_beta_20_100}
\end{figure}

We give the confidence-level vs. width and coverage plots for the setting presented in \autoref{sec:clt_ci} but where our data does not come from \(\theta \sim \betadist(1,1) = \text{Uniform(0,1)}\).
Specifically we consider three alternative true priors: \(\betadist(10,10)\), \(\theta \sim \betadist(100,20)\) and \(\theta \sim \betadist(20,100)\), the probability density functions of which are shown in \autoref{fig:beta-pdfs}.
These three settings are presented in \autoref{fig:iid_intervals_beta_10_10}, \autoref{fig:iid_intervals_beta_100_20}, and \autoref{fig:iid_intervals_beta_20_100} respectively.
As in \autoref{sec:clt_ci}, results are averaged over 100 values of $\theta$ drawn the given prior, each used to generate 200 random datasets.

We can see that the mismatched prior does affect the coverage of each method, in particular leading to Bayesian credible intervals that are too wide. 
However, as the amount of data increases, this problem resolves fairly quickly.

\begin{figure}[htbp]
    \centering
    \includegraphics[width=0.9\linewidth]{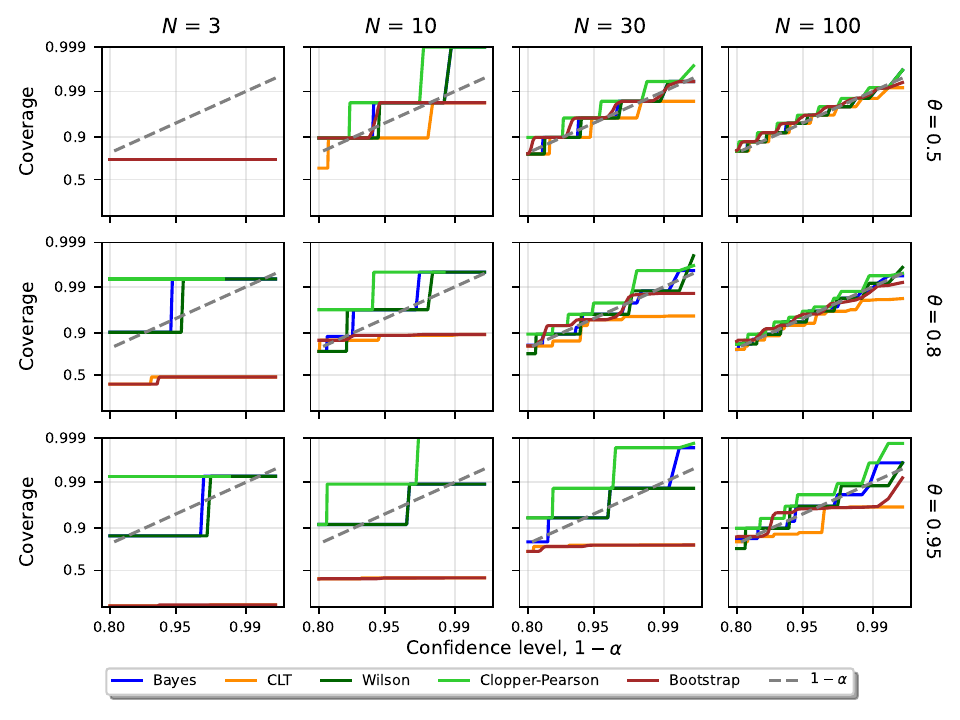}
    \vspace{-15pt}
    \caption{
       \textbf{IID question setting.} Coverage of intervals on \(\theta\) generated with fixed values of \(\theta \in \{0.5, 0.8, 0.95\}\).
    }
    \label{fig:iid_intervals_fixed_theta}
\end{figure}

In \autoref{fig:iid_intervals_fixed_theta}, we also present plots for experiments with fixed $\theta$ values, specifically with $\theta \in \{0.5, 0.8, 0.95\}$.
With each fixed $\theta$ value, we average results over 3000 simulated datasets.
These show much the same behaviour as we saw in Figs.~\ref{fig:iid_intervals_beta_10_10}-\ref{fig:iid_intervals_beta_20_100}, but are useful in that they represent a more frequentist approach to the evaluations, as opposed to the typical Bayesian setting in which we treat the `true' \(\theta\) as random rather than fixed.

\subsection{Clustered Questions Setting}\label{app:clustered_mismatch}
\begin{figure}[htbp]
    \centering
    \includegraphics[width=0.9\linewidth]{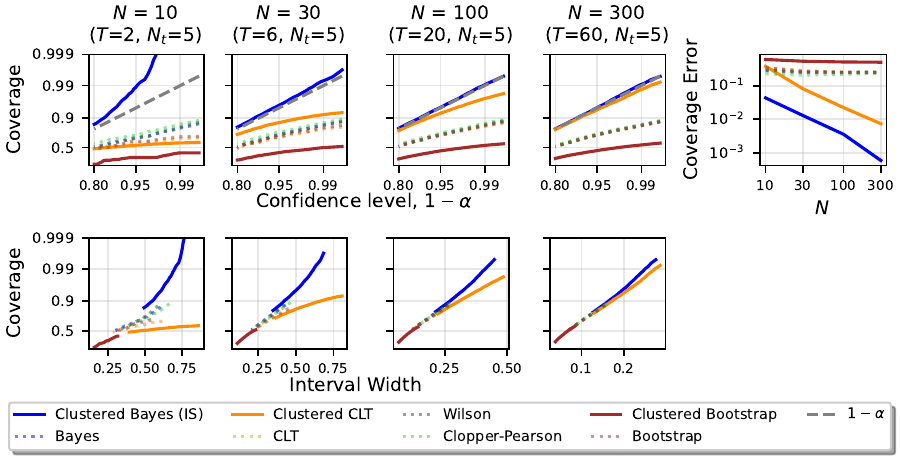}
    \caption{
       \textbf{Clustered question setting.} Coverage of intervals on \(\theta\) with mismatched \(\theta \sim \betadist(10,10)\) prior.
    }
    \label{fig:clustered_intervals_beta_10_10}
\end{figure}

\begin{figure}[htbp]
    \centering
    \includegraphics[width=0.9\linewidth]{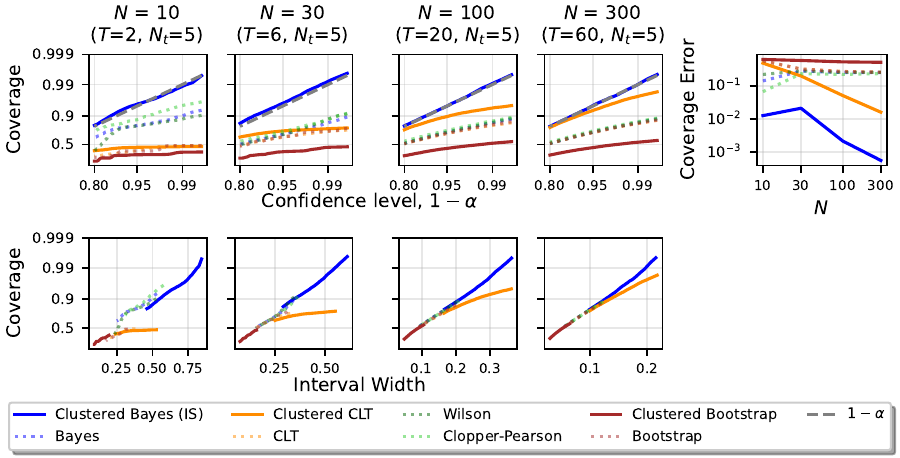}
    \vspace{-8pt}
    \caption{
        \textbf{Clustered question setting.} Coverage of intervals on \(\theta\) with mismatched \(\theta \sim \betadist(100,20)\) prior.
    }
    \label{fig:clustered_intervals_beta_100_20}
\end{figure}

\begin{figure}[htbp]
    \centering
    \includegraphics[width=0.9\linewidth]{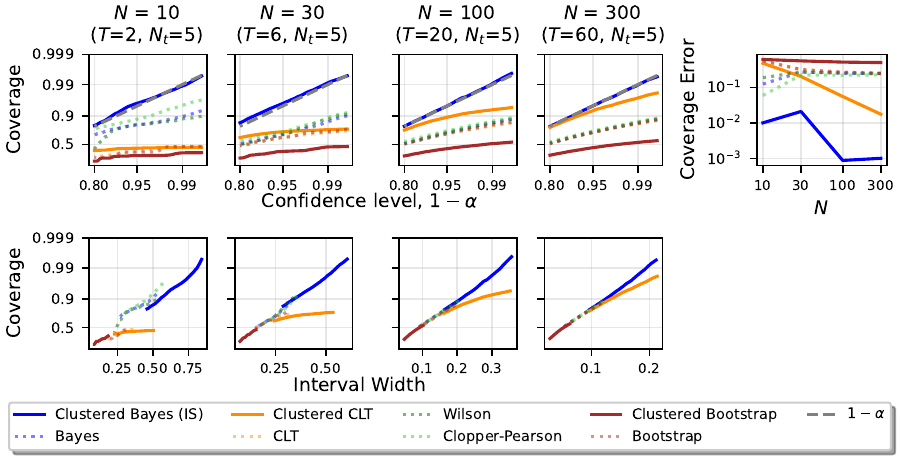}
    \vspace{-8pt}
    \caption{
        \textbf{Clustered question setting.} Coverage of intervals on \(\theta\) with mismatched \(\theta \sim \betadist(20,100)\) prior.
    }
    \label{fig:clustered_intervals_beta_20_100}
\end{figure}

Similarly to the previous section, we show results in the clustered questions setting (\autoref{sec:clt-clustered}) but with a mismatched true prior. 
\autoref{fig:clustered_intervals_beta_10_10} shows the results for \(\theta \sim \betadist(10,10)\), \autoref{fig:clustered_intervals_beta_100_20} shows the results for \(\theta \sim \betadist(100,20)\), and \autoref{fig:clustered_intervals_beta_20_100} shows the results for \(\theta \sim \betadist(20,100)\).
As in \autoref{sec:clt-clustered}, results are averaged over 100 values of $\theta$ drawn the given prior, each used to generate 200 random datasets.

In each case, we observe close to ideal performance from the clustered Bayes credible intervals, whereas all other methods tend to struggle to match their nominal coverage for small $N$.

\begin{figure}[htbp]
    \centering
    \includegraphics[width=0.90\linewidth]{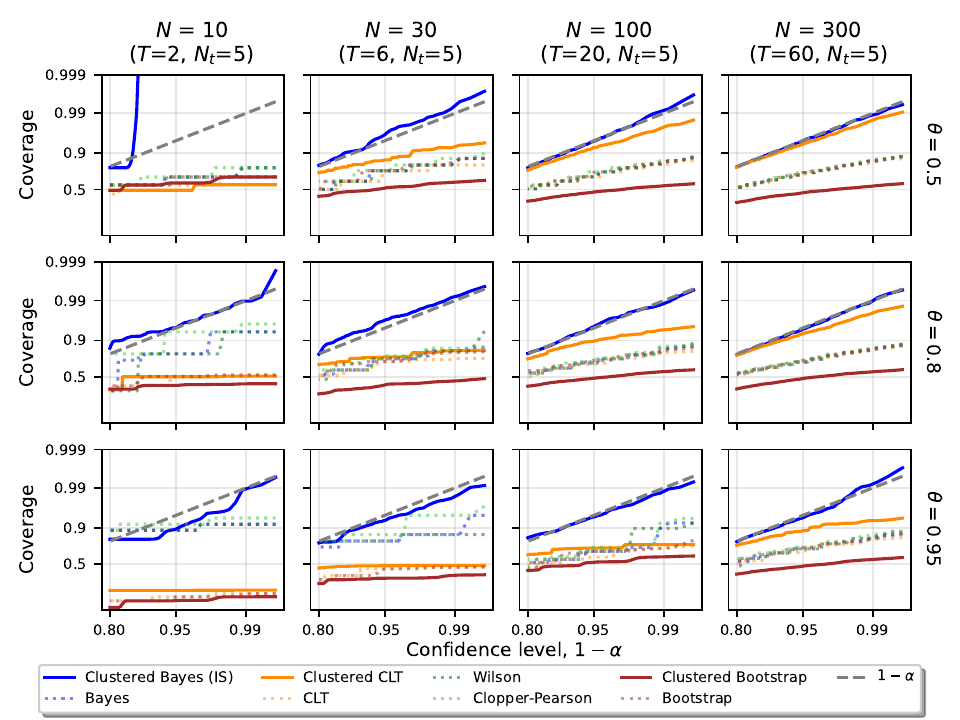}
    \vspace{-8pt}
    \caption{
        \textbf{Clustered question setting.} Coverage of intervals on \(\theta\) generated with fixed values of \(\theta \in \{0.5, 0.8, 0.95\}\).
    }
    \label{fig:clustered_intervals_fixed_theta}
\end{figure}

Again, we present results generated from fixed values of $\theta \in \{0.5, 0.8, 0.95\}$ rather than from a Bayesian prior.
With each fixed $\theta$ value, we average results over 3000 simulated datasets (each with a different sampled dispersion parameter $d \sim \gammadist(1,1)$.
These can be seen in \autoref{fig:clustered_intervals_fixed_theta} and show much the same behaviour as the previous plots.

\clearpage
\subsection{Independent Model Comparison}\label{app:indep_model_comparison}

Similarly to \autoref{app:iid_mismatch} and \autoref{app:clustered_mismatch}, we show results for the independent model comparison setting (\autoref{sec:indep_model_comparison}) but with a mismatched true prior, i.e. the actual data generating process does not match the prior that we assume for our the Bayesian model.

\begin{figure}[htbp]
    \centering
    \begin{subfigure}[b]{0.90\linewidth}
        \centering
        \includegraphics[width=\linewidth]{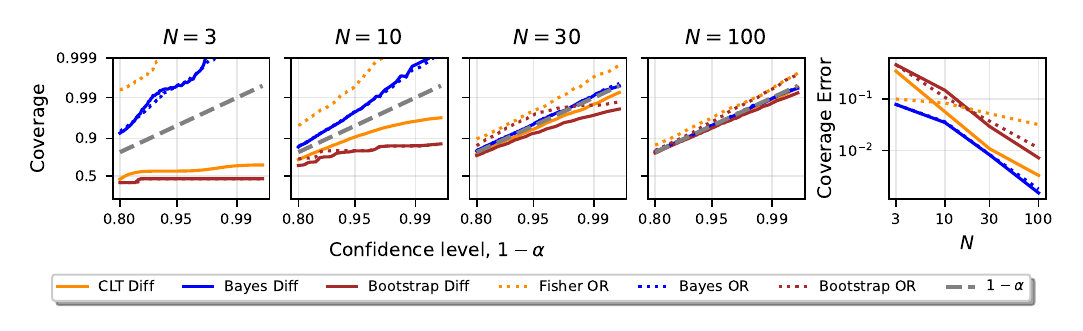}
        \caption{Prior used to generate data: \(\theta_A, \theta_B \sim \betadist(100,20)\)}
        \label{fig:mismatch_indep_comp_AB_a}
    \end{subfigure}%
    \vspace{1em}
    \begin{subfigure}[b]{0.90\linewidth}
        \centering
        \includegraphics[width=\linewidth]{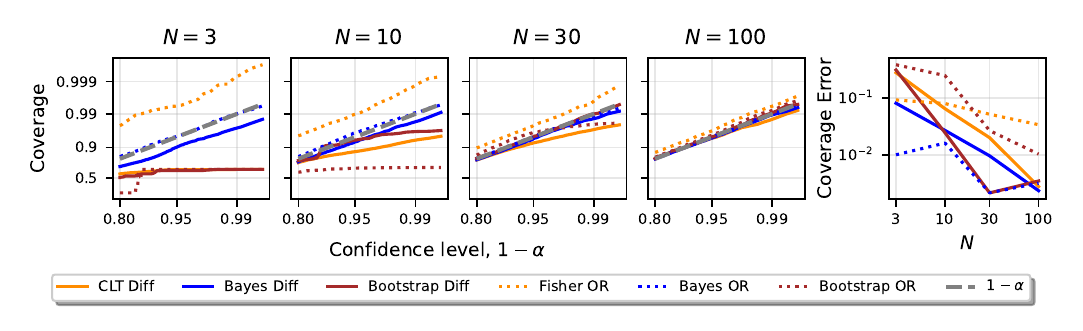}
        \caption{Prior used to generate data: \(\theta_A \sim \betadist(100,20)\) and \(\theta_B \sim \betadist(20,100)\).}
        \label{fig:mismatch_indep_comp_AB_b}
    \end{subfigure}
    % \vspace{-15pt}
    \caption{\textbf{Independent model comparison.} Prior mismatch: we use uniform, $\betadist(1,1)$, prior in the Bayesian model used to construct confidence intervals, whilst at test time the data is generated using a different prior. }
    \label{fig:indep_model_comparison_prior_mismatch_AandB}
\end{figure}

\begin{figure}[htbp]
    \centering
    \begin{subfigure}[b]{0.90\linewidth}
        \centering
        \includegraphics[width=\linewidth]{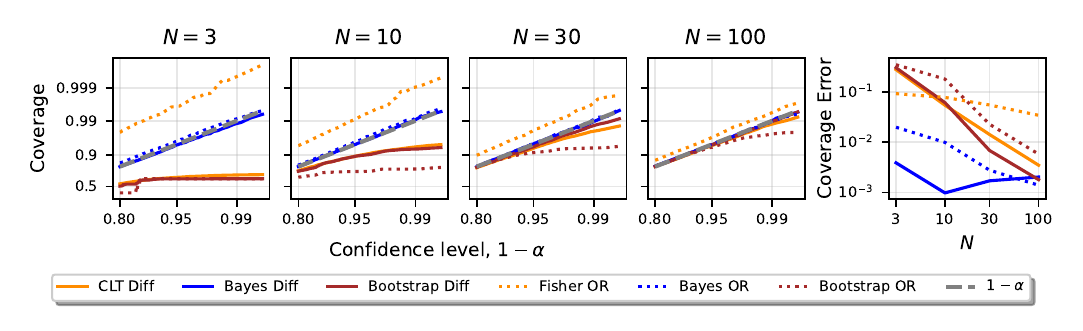}
        \vspace{-15pt}
        \caption{Prior used to generate data: \(\theta_B \sim \betadist(100,20)\)}
        \label{fig:mismatch_indep_comp_B_a}
    \end{subfigure}%
    \vspace{1em}
    \begin{subfigure}[b]{0.90\linewidth}
        \centering
        \includegraphics[width=\linewidth]{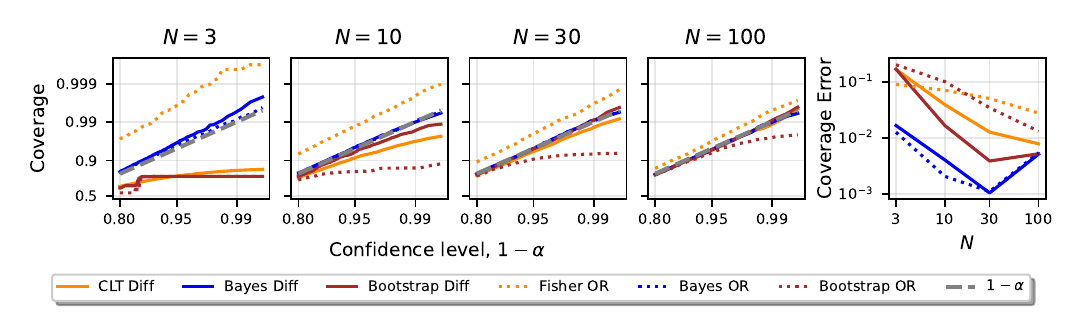}
        \vspace{-15pt}
        \caption{Prior used to generate data:  \(\theta_B \sim \betadist(10,10)\)}
        \label{fig:mismatch_indep_comp_B_b}
    \end{subfigure}%
    \vspace{1em}
    \begin{subfigure}[b]{0.90\linewidth}
        \centering
        \includegraphics[width=\linewidth]{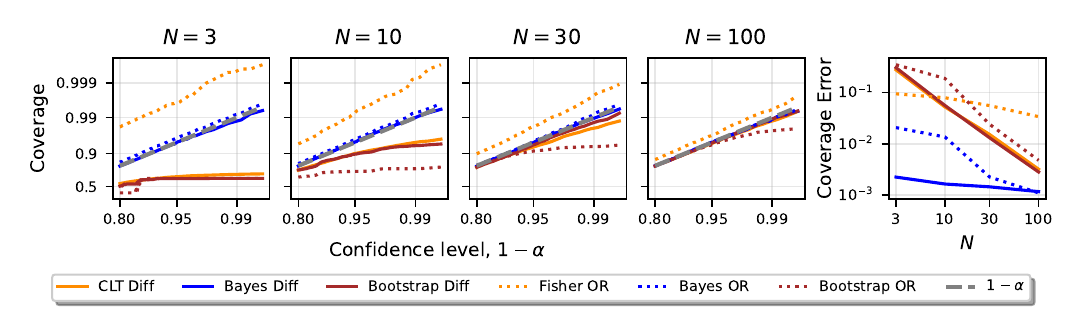}
        \vspace{-15pt}
        \caption{Prior used to generate data: \(\theta_B \sim \betadist(20,100)\)}
        \label{fig:mismatch_indep_comp_B_c}
    \end{subfigure}
    % \vspace{-15pt}
    \caption{\textbf{Independent model comparison.} Prior mismatch: we use uniform, $\betadist(1,1)$, prior in the Bayesian model used to construct confidence intervals.
    At test time, the accuracy of model A is sampled from $\betadist(1, 1)$ whilst that of model B is sampled from a different prior.}
    \label{fig:indep_model_comparison_prior_mismatch_B}
\end{figure}

\begin{figure}[htbp]
    \centering
    \includegraphics[width=0.90\linewidth]{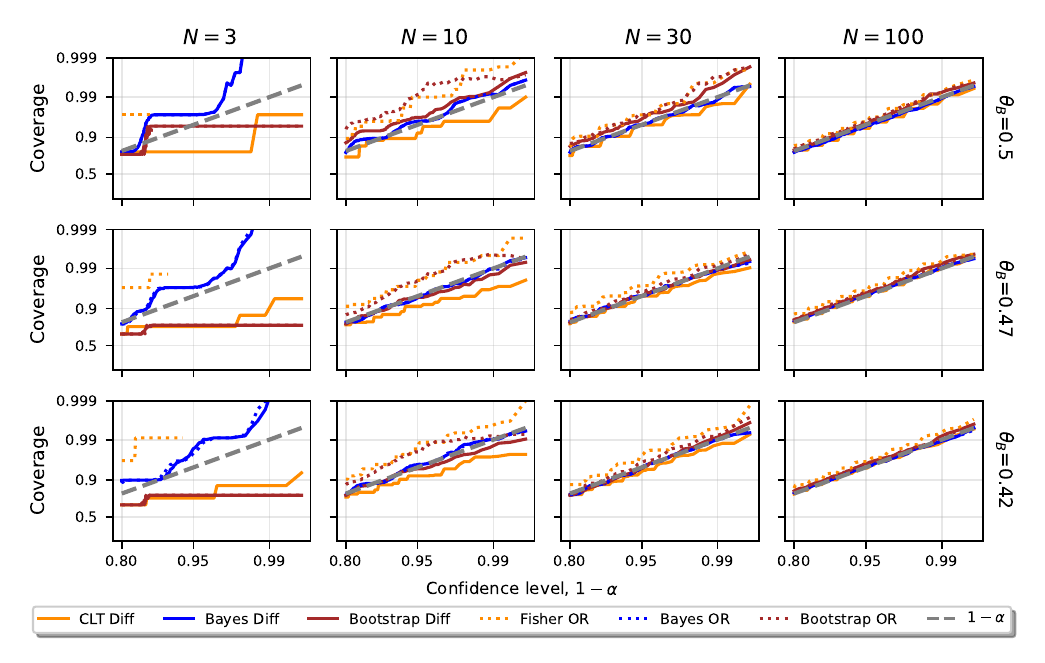}
    \vspace{-15pt}
    \caption{
        \textbf{Independent model comparison.} Success probabilities are fixed at $\theta_A = 0.5$ and $\theta_B \in \{0.5, 0.47, 0.42\}$.
    }
    \label{fig:indep_model_comparison_thetaA_0.5}
\end{figure}

\begin{figure}
    \centering
    \includegraphics[width=0.90\linewidth]{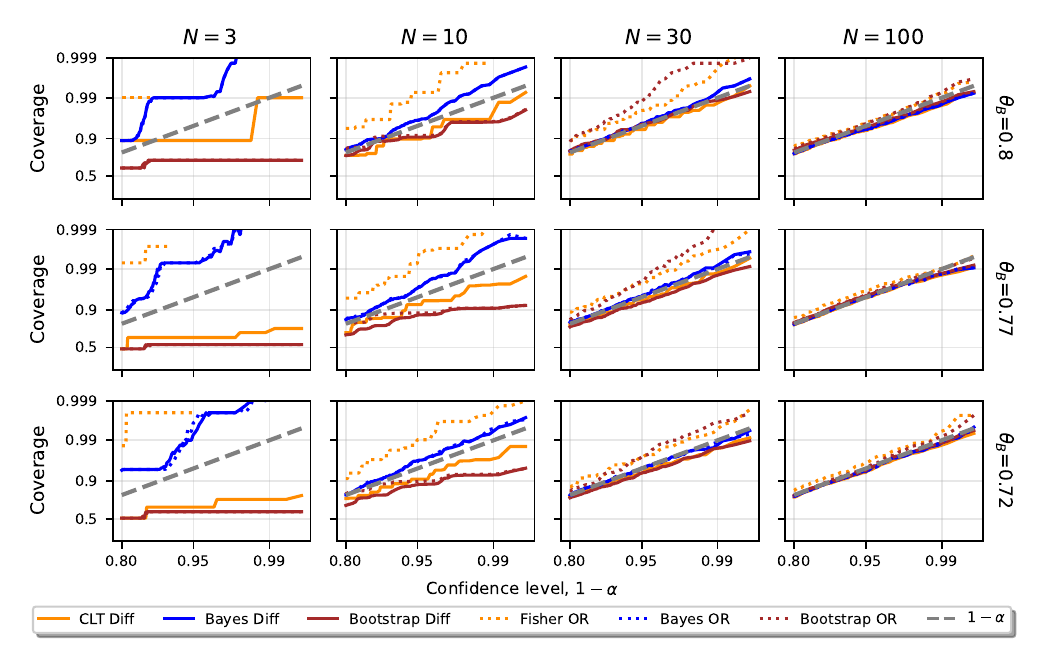}
    \vspace{-15pt}
    \caption{
        \textbf{Independent model comparison.} Success probabilities are fixed at $\theta_A = 0.8$ and $\theta_B \in \{0.8, 0.77, 0.72\}$.
    }
    \label{fig:indep_model_comparison_thetaA_0.8}
\end{figure}

\begin{figure}
    \centering
    \includegraphics[width=0.90\linewidth]{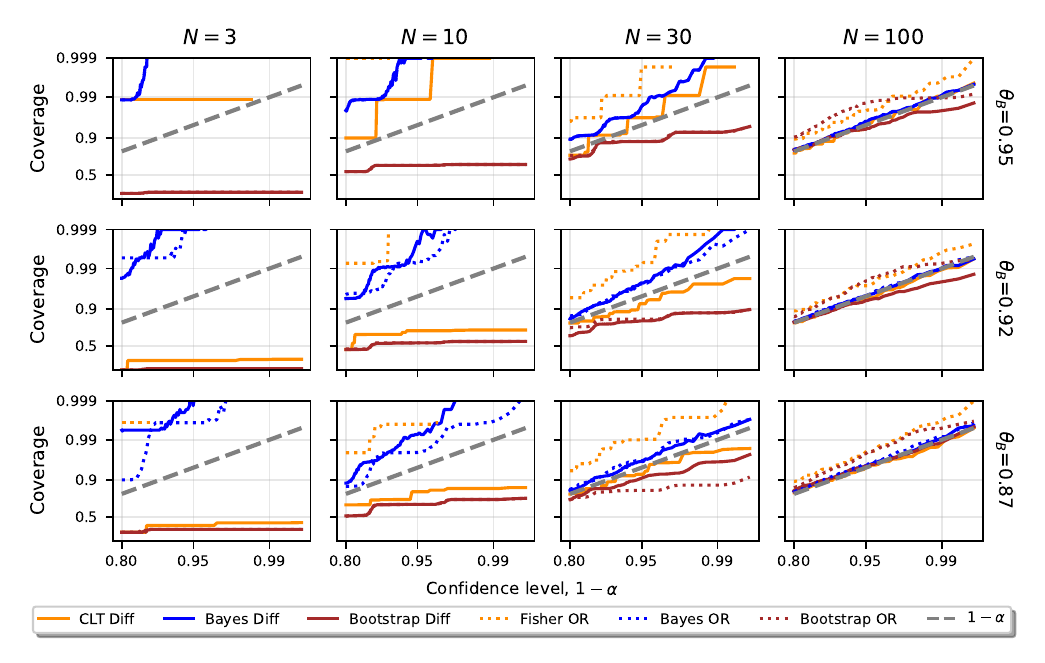}
    \vspace{-15pt}
    \caption{
        \textbf{Independent model comparison.} Success probabilities are fixed at $\theta_A = 0.95$ and $\theta_B \in \{0.95, 0.92, 0.87\}$.
    }
    \label{fig:indep_model_comparison_thetaA_0.95}
\end{figure}

% Unless otherwise stated, we use the same simulation framework as in the main paper. We consider the following settings:
Specifically, we consider the following settings:
\begin{itemize}
    \item Neither model A nor model B has a uniform prior: \autoref{fig:mismatch_indep_comp_AB_a}  shows results in which both \(\theta_A, \theta_B \sim \betadist(100,20)\), whilst \autoref{fig:mismatch_indep_comp_AB_b} presents results in which we have \(\theta_A \sim \betadist(100,20)\) and \(\theta_B \sim \betadist(20,100)\)
    \item  Keep the same prior for \(\theta_A\), that is, \(\betadist(1, 1) = \uniform[0,1]\) and vary the prior for \(\theta_B\) between \(\betadist(100,20)\) (\autoref{fig:mismatch_indep_comp_B_a}); \(\betadist(10,10)\) (\autoref{fig:mismatch_indep_comp_B_b}); and \(\betadist(20,100)\) (\autoref{fig:mismatch_indep_comp_B_c}).
    \item Fixed values for \(\theta_A\)and \(\theta_B\) as follows: \(\theta_A \in \{0.5, 0.8, 0.9\}\) with \(\theta_B\) taking values in \(\{\theta_A, \theta_A - 0.3, \theta_A-0.8\}\) in each setting. These are shown in \autoref{fig:indep_model_comparison_thetaA_0.5}, \autoref{fig:indep_model_comparison_thetaA_0.8}, and \autoref{fig:indep_model_comparison_thetaA_0.95} respectively.
    Results are shown averaged over 3000 simulated datasets for each $(\theta_A, \theta_B)$ pair.
\end{itemize}

\clearpage
\subsection{Paired Questions Setting}\label{app:paired_mismatch}
\begin{figure}[htbp]
    \centering
    \includegraphics[width=0.90\linewidth]{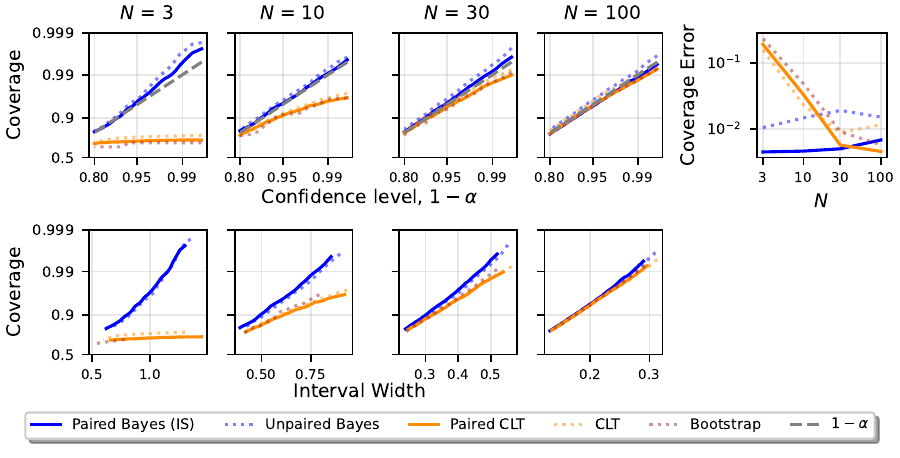}
    \vspace{-15pt}
    \caption{
        \textbf{Paired question model comparison setting.} Coverage of intervals on $\theta_A - \theta_B$ with mismatched \(\theta_B \sim \betadist(10,10)\) prior and \(\theta_A \sim \betadist(1,1)\).}
    \label{fig:paired_intervals_beta_10_10}
\end{figure}

\begin{figure}[htbp]
    \centering
    \includegraphics[width=0.90\linewidth]{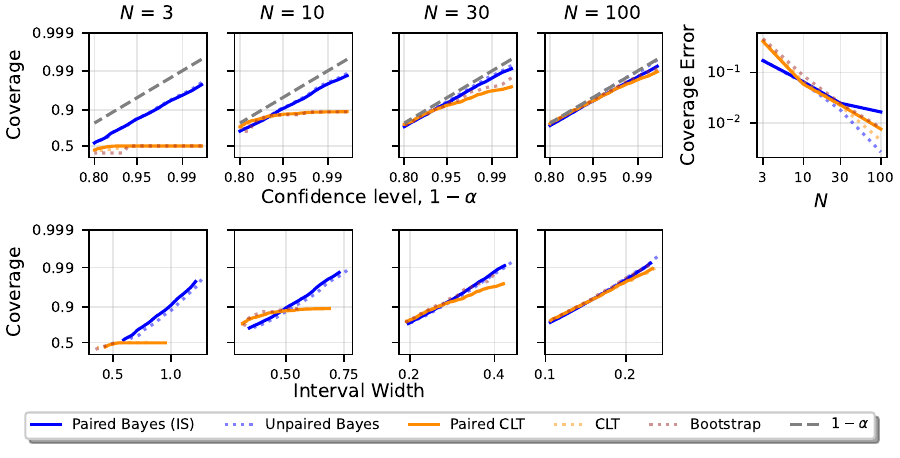}
    \vspace{-15pt}
    \caption{
        \textbf{Paired question model comparison  setting.} Coverage of intervals on $\theta_A - \theta_B$ with mismatched \(\theta_B \sim \betadist(100,20)\) prior and \(\theta_A \sim \betadist(1,1)\))
    }
    \label{fig:paired_intervals_beta_100_20}
\end{figure}

\begin{figure}[htbp]
    \centering
    \includegraphics[width=0.9\linewidth]{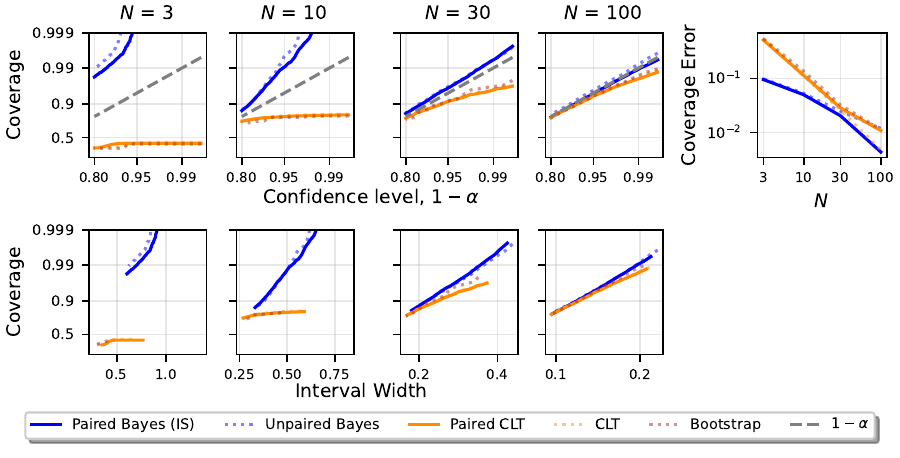}
    \vspace{-15pt}
    \caption{
        \textbf{Paired question model comparison  setting.} Coverage of intervals on $\theta_A - \theta_B$ with mismatched \(\theta_B \sim \betadist(20,100)\) prior and \(\theta_A \sim \betadist(1,1)\).
    }
    \label{fig:paired_intervals_beta_20_100}
\end{figure}

\begin{figure}[htbp]
    \centering
    \includegraphics[width=0.9\linewidth]{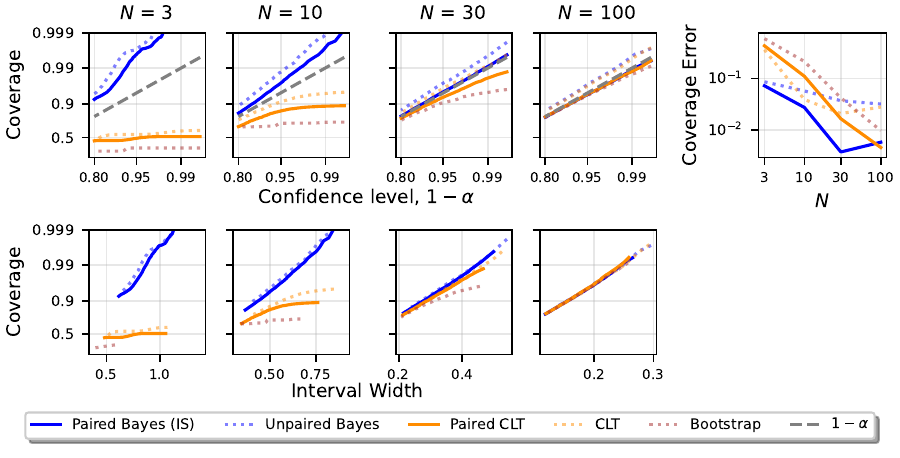}
    \vspace{-15pt}
    \caption{
        \textbf{Paired question model comparison  setting.} Coverage of intervals on $\theta_A - \theta_B$ with mismatched \(\theta_A, \theta_B \sim \betadist(20,100)\) priors.
    }
    \label{fig:paired_intervals_both_beta_100_20}
\end{figure}

\begin{figure}[htbp]
    \centering
    \includegraphics[width=0.9\linewidth]{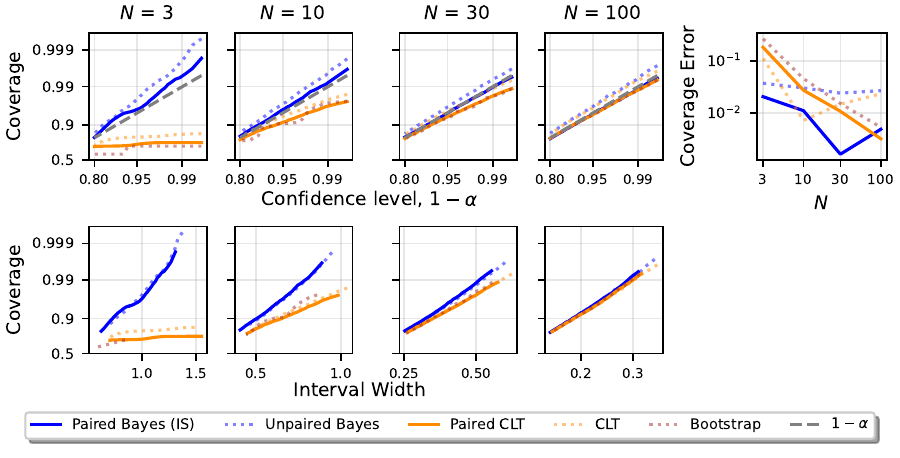}
    \vspace{-15pt}
    \caption{
        \textbf{Paired question model comparison  setting.} Coverage of intervals on $\theta_A - \theta_B$ with mismatched priors: \(\theta_A \sim \betadist(100,20)\), and \(\theta_B \sim \betadist(20,100)\).
    }
    \label{fig:paired_intervals_both_mismatched_differently}
\end{figure}

Similarly to \autoref{app:iid_mismatch} and \autoref{app:clustered_mismatch}, we show results for the paired questions setting (\autoref{sec:paired}) but with a mismatched true prior. 
In particular, we keep the same prior for \(\theta_A\) (that is, \(\betadist(1,1) = \uniform[0,1]\)) but vary the prior for \(\theta_B\) between \(\betadist(10,10)\) (\autoref{fig:paired_intervals_beta_10_10}); \(\betadist(100,20)\) (\autoref{fig:paired_intervals_beta_100_20}); and \(\betadist(20,100)\) (\autoref{fig:paired_intervals_beta_20_100}).
We also present results where neither model A nor model B has a uniform prior: \autoref{fig:paired_intervals_both_beta_100_20}  shows results in which both \(\theta_A, \theta_B \sim \betadist(100,20)\), whilst \autoref{fig:paired_intervals_both_mismatched_differently} presents results in which we have \(\theta_A \sim \betadist(100,20)\) and \(\theta_B \sim \betadist(20,100)\).
% In \autoref{fig:paired_intervals_large_rho}, we present results in which $\theta_A, \theta_B \sim \betadist(1,1)$ as usual, but here our correlation coefficient is sampled as $\rho \sim \uniform[0.33,1]$.
% This is because one would expect to find high correlation in true paired eval results.
As in \autoref{sec:clt_ci}, results are averaged over 100 values of $(\theta_A, \theta_B)$ drawn the specified priors, with each $(\theta_A, \theta_B)$ pair used to generate 200 random datasets.

In each case, we observe the coverage of Bayesian credible intervals approaching the ideal $1-\alpha$ line at least as quickly (in terms of an increasing $N$) as the paired-CLT confidence intervals.
We also observe the paired Bayesian credible intervals generally outperforming the unpaired Bayesian intervals.

% \begin{figure}[htbp]
%     \centering
%     \includegraphics[width=0.9\linewidth]{fig/exp4-4_pos_rho.pdf}
%     \vspace{-15pt}
%     \caption{
%         \textbf{Paired question model comparison  setting.} Coverage of intervals on $\theta_A - \theta_B$ with mismatched \(\rho \sim \uniform[0.33,1]\) priors.
%     }
%     \label{fig:paired_intervals_large_rho}
% \end{figure}

\begin{figure}[htbp]
    \centering
    \includegraphics[width=0.82\linewidth]{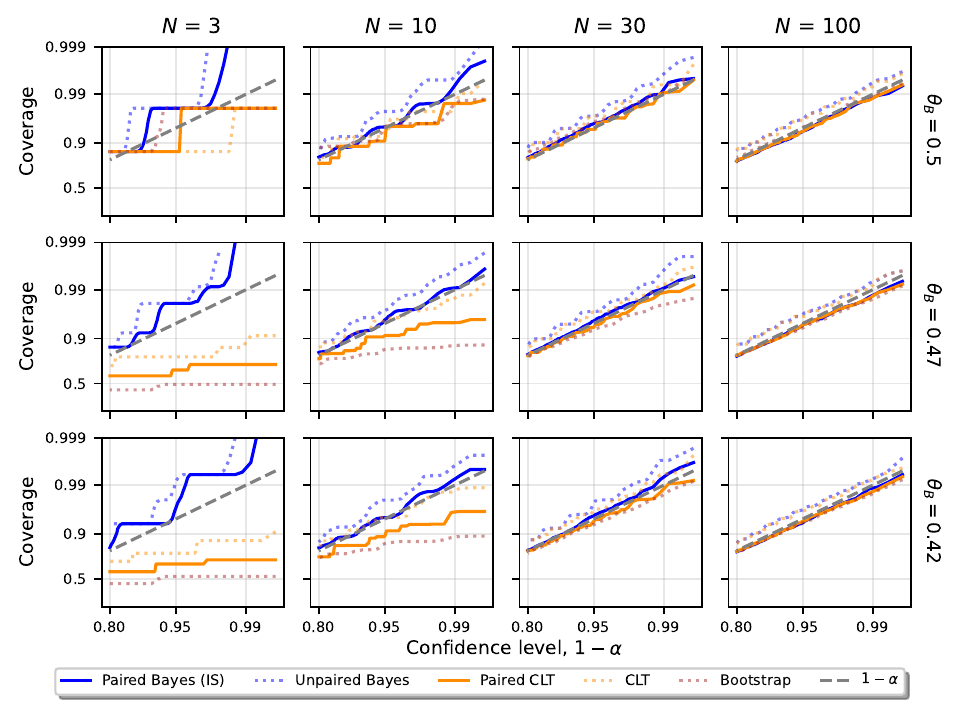}
    \vspace{-15pt}
    \caption{
        \textbf{Paired question model comparison setting.} Coverage of intervals on $\theta_A - \theta_B$ generated with \(\theta_A = 0.5\) and \(\theta_B \in \{0.5, 0.47, 0.42\}\).
    }
    \label{fig:paired_intervals_fixed_theta_0.5}
\end{figure}

\begin{figure}[htbp]
    \centering
    \includegraphics[width=0.82\linewidth]{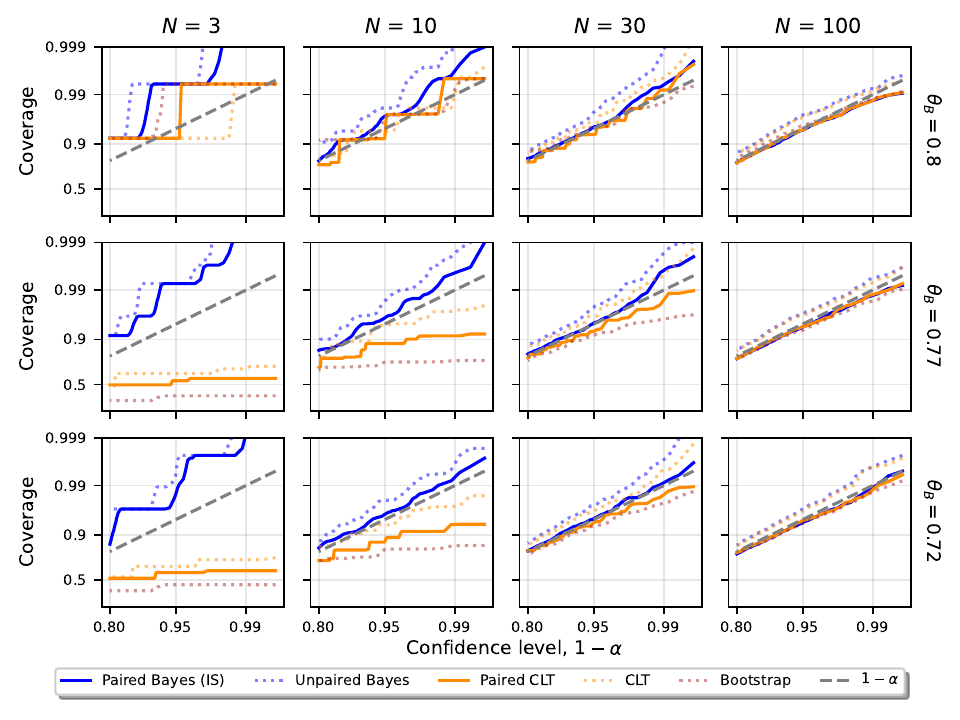}
    \vspace{-15pt}
    \caption{
        \textbf{Paired question model comparison setting.} Coverage of intervals on $\theta_A - \theta_B$ with \(\theta_A = 0.8\) and \(\theta_B \in \{0.8, 0.77, 0.72\}\).
    }
    \label{fig:paired_intervals_fixed_theta_0.8}
\end{figure}

\begin{figure}[htbp]
    \centering
    \includegraphics[width=0.82\linewidth]{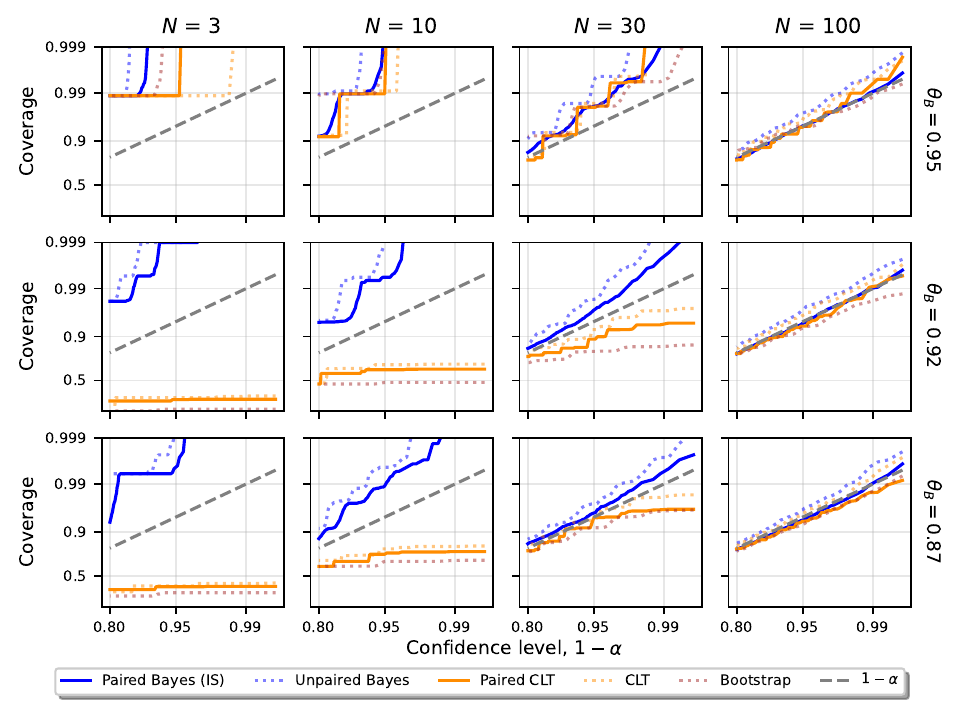}
    \vspace{-15pt}
    \caption{
        \textbf{Paired question model comparison  setting.} Coverage of intervals on $\theta_A - \theta_B$ with \(\theta_A = 0.95\) and \(\theta_B \in \{0.95, 0.92, 0.87\}\).
    }
    \label{fig:paired_intervals_fixed_theta_0.95}
\end{figure}

Finally, for this section, we also present results with the same fixed values for \(\theta_A\)and \(\theta_B\) as in \autoref{app:indep_model_comparison}: \(\theta_A \in \{0.5, 0.8, 0.9\}\) with \(\theta_B\) taking values in \(\{\theta_A, \theta_A - 0.3, \theta_A-0.8\}\) in each setting.
Results are shown averaged over 3000 simulated datasets for each $(\theta_A, \theta_B)$ pair.
These are shown in \autoref{fig:paired_intervals_fixed_theta_0.5}, \autoref{fig:paired_intervals_fixed_theta_0.8}, and \autoref{fig:paired_intervals_fixed_theta_0.95} respectively.

Whilst the coverage of both the Bayesian credible intervals and the paired CLT-based confidence intervals are very similar when $\theta_A = \theta_B$, we see much more robust behaviour from the Bayesian method when we increase the difference between $\theta_A$ and $\theta_B$.

\FloatBarrier

% \vspace{-25pt}
\section{Error Bars on the $F_1$ Score} \label{app:f_scores_ablations}

% \vspace{-23pt}
% \subsection{Interval width}
\begin{figure*}[th]
    \centering
    % \vspace{-15pt}
    \includegraphics[width=0.65\linewidth]{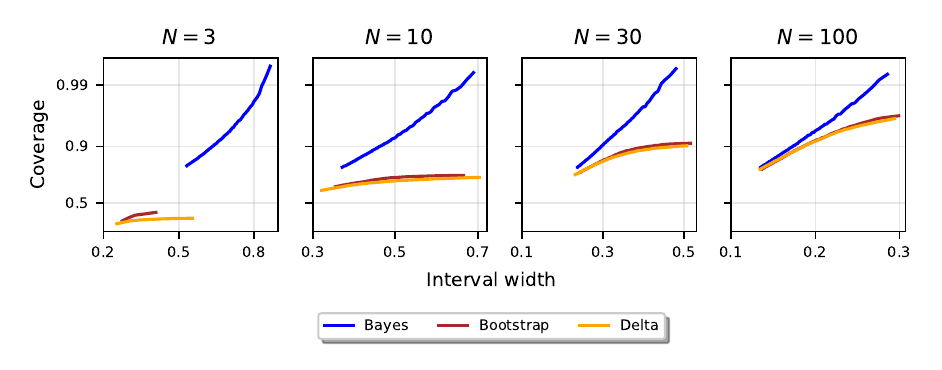}
    \vspace{-15pt}
    \caption{\textbf{Error bars for the $F_1$-score}. Coverage vs. interval-width for various interval-calculation methods.
    }
    \label{fig:f_score_size}
\end{figure*}
\vspace{-15pt}
% \subsection{HDI vs QBI Ablation} \label{app:hdi_vs_qbi}
\begin{figure}[h!]
    \centering
    \includegraphics[width=0.8\linewidth]{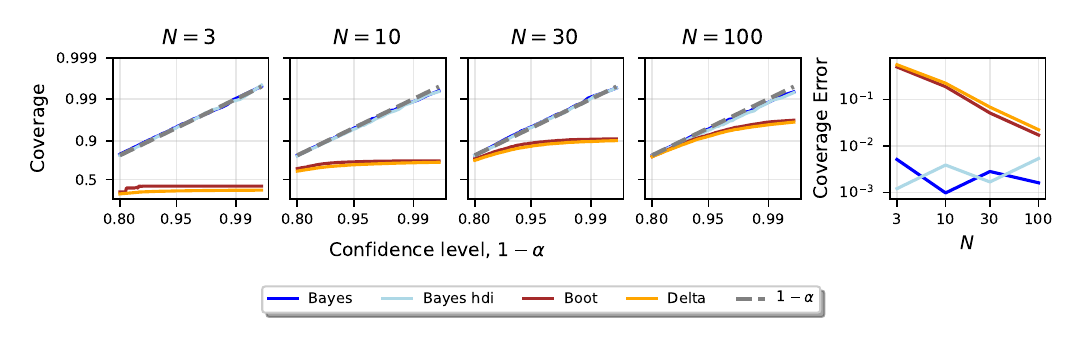}
    \vspace{-15pt}
    \caption{
        \textbf{Error bars for the $F_1$-score}. 
    Coverage vs. confidence level for Bayesian and bootstrap intervals on $F_1$ scores using highest posterior density interval (HDI) instead of quantile-based interval (QBI), which was presented in \autoref{fig:f_scores}.
    }
    \label{fig:f_scores_hdi}
\end{figure}

\end{document}